\documentclass[accepted]{melba}

\usepackage{mwe} 
\usepackage{amsmath,amsfonts,amssymb}
\usepackage{graphicx}
\usepackage{setspace}
\usepackage{tocloft}
\usepackage{lineno}
\usepackage{booktabs,multirow,adjustbox,siunitx}
\usepackage{amsmath,amsfonts}

\melbaid{2026:052}
\doi{10.59275/j.melba.2026-3893}
\melbaauthors{Zhenhao Guo et al}  
\email{rud4004@med.cornell.edu}
\volume{2026}
\firstpageno{803}
\melbayear{2026}
\datesubmitted{2025-11-14}
\datepublished{2026-10-09}  

\ShortHeadings{From Classification to Cross-Modal Understanding}{Guo}

\title{From Classification to Cross-Modal Understanding: Leveraging Vision-Language Models for Fine-Grained Renal Pathology}

\author{
  \firstname Zhenhao \surname Guo\aff{1},
  \name Rachit Saluja\aff{2},
  \name Tianyu Shi\aff{3},
  \name Tianyuan Yao\aff{4},
  \name Quan Liu\aff{4},
  \name Junchao Zhu\aff{4},
  \name Haibo Wang\aff{5},
  \name Daniel Reisenbüchler\aff{6},
  \name Benjamin Liechty\aff{7},
  \name David J. Pisapia\aff{7},
  \name Kenji Ikemura\aff{7},
  \name Steven Salvatoree\aff{7},
  \name Surya Seshane\aff{7},
  \name Mert R. Sabuncu\aff{2,7},
  \name Yihe Yang\aff{7,8},
  \name Ruining Deng\aff{4,7}
}

\affiliations{%
  \num 1 \addr New York University, New York, NY 10012, USA \\
  \num 2 \addr Cornell Tech, New York, NY 10044, USA \\
  \num 3 \addr Sichuan University, Chengdu, 610207, CN \\
  \num 4 \addr Vanderbilt University, Nashville, TN, 37235, USA \\
  \num 5 \addr Carnegie Mellon University, Pittsburgh, PA 15213, USA \\
  \num 6 \addr University of Regensburg, Regensburg, Bavaria 93053, DE \\
  \num 7 \addr Weill Cornell Medicine, New York, NY 10065, USA \\
  \num 8 \addr Northwell Health, New Hyde Park, NY 11040, USA
}

\abstract{
Fine-grained glomerular subtyping is essential for kidney biopsy interpretation, yet it presents a harder setting than the broad organ- or disease-level recognition tasks commonly used in vision-language model (VLM) evaluation. In this task, all categories arise within the same renal microanatomic unit, and clinically meaningful subtypes often differ by subtle, overlapping, and continuum-like morphological cues. Clinically valuable labels are also scarce, making fully supervised image-only paradigms difficult to translate to practice. Although VLMs are promising for data-efficient pathology learning, it remains unclear how their cross-modal representations should be adapted and evaluated for such within-structure discrimination under realistic few-shot constraints. To our knowledge, this study establishes the first fine-grained renal pathology benchmark for few-shot VLM adaptation and systematically examines how supervision level, model design, and adaptation strategy shape both classification performance and multimodal representation structure. Beyond accuracy, area under the receiver operating characteristic curve (AUC), F1 score, image-text alignment, positive-negative similarity separation, embedding geometry, and confidence calibration quantified by expected calibration error (ECE), we ask when performance gains reflect more reliable cross-modal understanding. Across renal pathology cohorts, additional supervision consistently improves discrimination, but reliable adaptation is not explained by alignment alone; stronger performance is associated with larger class separation, more coherent embedding organization, and better awareness of confidence-related failure modes under domain variation. Within this retrospective two-cohort evaluation, these findings move assessment beyond coarse classification and accuracy alone and provide comparative evidence to inform model selection and future translational validation. Our implementation has now been publicly released for reproducibility on GitHub at \url{https://github.com/ddrrnn123/glo-vlms}.
}

\keywords{Vision-Language Model, Fine-Grained Classification, Fine-tuning, Few-shot Learning, Digital Renal Pathology}

\begin{document}

\twocolumn[\maketitle]

\section{Introduction}
	\enluminure{R}{enal} \emph{histopathology} is central to diagnosis, subtyping, and prognosis in kidney disease~\cite{feng2023artificial,sethi2019standardized,schnuelle2023renal,wilson2022whole}. Fine-grained glomerular subtypes carry therapeutic and risk-stratification implications beyond coarse lesion detection~\cite{of2009oxford,rafieian2013significance,chen2022retrospective,lu2025clinical,deng2022cross,yu2025glo}. Yet high-quality labels are expensive, inter-expert agreement is limited, and cross-institution variation in staining and scanners induces domain shift~\cite{duenweg2023whole}. In nephrectomy cohorts of renal cell carcinoma (RCC), surgical clamping of the renal artery, veins, or ureter produces ischemic artifacts such as glomerular basement membrane (GBM) wrinkling and capsular distention, so many glomeruli share similar ischemic-appearing changes. RCC patients also share common risk factors with diabetes, hypertension, and metabolic syndrome, and many glomeruli lie along a continuum from ischemia to segmental and then global sclerosis, often showing GBM wrinkling, multilayering of Bowman's capsule, and tuft-capsule adhesions in the same glomerulus. These surgical and clinical factors blur the boundaries between ischemic, segmentally sclerotic, and globally sclerotic glomeruli, making it difficult for algorithms to distinguish between these groups. In practice, small labeled cohorts are the norm, and clinicians often reason with a few confirmed exemplars as references~\cite{crowley2003development}. Recent advances in foundation VLMs have transformed computational pathology, offering generalizable representations that bridge visual and textual modalities~\cite{cui2022survival,chanda2024new,bilal2025foundation,ochi2025pathology}. However, standardized evidence for their fine-grained adaptation remains limited, complicating principled selection across backbones and parameter-efficient fine-tuning strategies. Prior studies typically evaluate only a single task granularity~\cite{campanella2019clinical}, supervision regime~\cite{lu2021improve}, or adaptation strategy at a time~\cite{yao2022self}, often focusing on coarse normal-vs-diseased discrimination~\cite{WANG2025109670}, training under full supervision~\cite{yu2025glo}, or image-only classification with one backbone and one split~\cite{yao2022self}, and rarely characterizing how performance scales with sample size~\cite{taher2025large,hosseinzadeh2021systematic}. At the same time, the degree to which visual and textual features are aligned is increasingly viewed as central to the robustness and clinical reliability of VLMs in pathology~\cite{chen2024survey,li2024multimodal,hanna2025future}. We address these gaps by (i) establishing a framework for fine-grained glomerular subtyping, (ii) characterizing model performance in a clinically realistic few-shot regime, and (iii) analyzing multimodal image-text feature alignment and between-class separability. We evaluate both general-purpose and pathology-specialized backbones across prevailing architectures and multiple parameter-efficient adaptation strategies across few-shot settings using Monte Carlo resampling. Beyond measuring performance with classification metrics, we perform a cross-modal embedding analysis: we quantify image-text feature alignment and between-class separability as a function of supervision level, task granularity, and adaptation strategy. We then relate these representational properties to model performance and ECE. Taken together, this reveals how supervision and adaptation jointly shape multimodal alignment and downstream behavior, providing practical guidance for model selection under realistic data constraints.

Our contributions are threefold:
\begin{enumerate}
\item To our knowledge, we establish the first VLM benchmark for fine-grained glomerular subtyping in renal pathology, targeting pathologically actionable distinctions beyond coarse disease classification.

\item We characterize model behavior under realistic data limitations by restricting supervision to a few labeled exemplars per class and analyzing how backbone, parameter-efficient adaptation, and sample size jointly affect performance in general-purpose and pathology-specialized foundation models.

\item We introduce embedding-space analyses that quantify multimodal image-text feature alignment and between-class separability, and we show that these metrics track downstream accuracy and confidence calibration, revealing trends that inform future cross-modal learning under few-shot constraints.
\end{enumerate}

\begin{figure*}[!t]
\centering
\includegraphics[width=0.99\textwidth]{{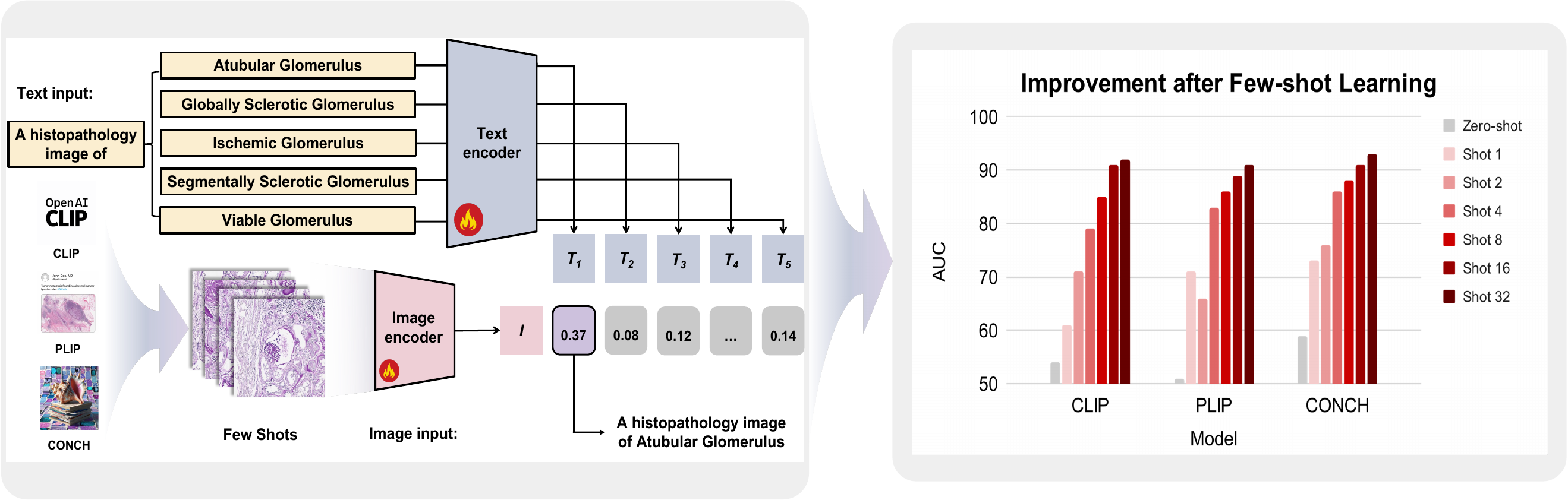}}
\caption{\textbf{Pipeline overview.} We train the image encoder and the text encoder in a few-shot setting, where only a small number of labels are provided for each glomerular subtype. For each shot, patches and their class descriptions are encoded, and the model is updated. As the number of shots increases, the performance improves.}
\label{fig1:Pipeline Overview}
\end{figure*}

\section{Related Works}
	\subsection{Few-shot learning for medical image classification}
Renal pathologists have established morphology-based systems for fine-grained glomerular classification. Recent work has further refined outcome assessment in focal segmental glomerulosclerosis through a new index~\cite{tervaert2010pathologic,bajema2018revision,chan2024new}. Together, these expert-defined schemas provide fine-grained and clinically meaningful labels that now serve as reliable ground truth for training AI models. However, pathology data are inherently scarce, costly to annotate, and highly heterogeneous across institutions and imaging protocols~\cite{madabhushi2016image,shi2019effects,li2025survey}. As a result, few-shot learning naturally emerges as a practical and clinically aligned approach that maintains robust performance with only a handful of examples per class. Prior studies indicate that such methods can substantially mitigate data scarcity and even approach fully supervised accuracy with very limited samples~\cite{pachetti2024systematic,shakeri2022fhist}. Accordingly, few-shot learning provides a compelling direction for developing reliable and data-efficient pathology AI systems under low-annotation constraints. However, prior work rarely reports shot-scaling behavior or evaluates adaptation under explicit cross-center domain shift, leaving unclear how shot budget affects robustness and which adaptation strategy remains reliable.

In our work, we treat fine-grained renal pathology classification as a clinically grounded few-shot medical image classification problem and disentangle the effects of three key factors: foundation model choice, fine-tuning strategy, and shot budget. Under controlled whole-slide image (WSI)-level splits and repeated runs, we systematically vary the backbone (general-purpose and pathology-specialized), the adaptation method (full vs parameter-efficient updates), and the number of labeled examples per class, and report shot-scaling behavior across these settings. This study is designed to provide practical insights into which combinations are most data-efficient and how additional annotations translate into measurable gains, informing model selection and annotation investment under realistic data constraints.

	\subsection{Fine-grained Glomerular Classification}
In recent years, deep learning and VLMs have achieved substantive progress in fine-grained glomerular classification. Deep learning, grounded in morphological patterns, has produced end-to-end convolutional pipelines on WSIs that cover glomerulus detection, disease categorization, and key lesion identification~\cite{yao2022self,lu2021improve,yu2025glo,weis2022assessment,lei2024artificial,li2025fine}. At the same time, the feasibility of VLMs for pathology classification has been increasingly demonstrated. These VLM-based methods have shown strong performance~\cite{zheng2025benchmarking,guo2025glo}. Taken together, these studies provide converging evidence that both deep learning and VLMs are feasible for fine-grained glomerular classification. However, most existing studies emphasize Top-1 performance, while practical deployment constraints and reliability measures are rarely reported in a unified and systematic manner. In particular, confidence calibration (e.g., ECE) and decision-margin behavior (e.g., boundary margin) are seldom quantified, limiting evidence about confidence reliability under clinical conditions. Consequently, it remains unclear how these methods translate into dependable and deployable decision support in renal pathology workflows.

To address these limitations, we perform a systematic evaluation of few-shot fine-grained glomerular classification under clinically realistic data constraints. In our work, we compare general-purpose and pathology-specialized backbones as well as multiple adaptation strategies under matched shot budgets and cross-center validation, and we report not only discrimination metrics (Accuracy, AUC, and F1) but also confidence calibration. Moreover, we measure ECE and boundary margin to characterize whether predicted probabilities are calibrated and whether decisions are made with a sufficient separation between competing classes. This combined view provides a broader research assessment of classification performance and confidence behavior under limited supervision and cross-center variation.

\subsection{Cross-modal Vision Language Models on Renal Pathology}
In clinical practice, physicians engage in a multimodal and integrative diagnostic process, synthesizing information from diverse sources. This cross-modal verification ensures that diagnostic reasoning is both reliable and interpretable~\cite{roth2021multispecialty,cui2024enhancing,jin2025anomaly}. Analogously, to endow artificial intelligence systems with comparable credibility in medical contexts, it is essential to move beyond unimodal classification paradigms and promote modality alignment. Such alignment enables the model to ground its predictions in coherent morphological and textual evidence, thereby enhancing transparency and interpretability~\cite{kline2022multimodal,simon2025future}. Moreover, effective modality alignment contributes to improved empirical performance, as it facilitates the extraction of stable and domain-general representations within a unified semantic space~\cite{wang2020understanding,xu2024style,wang2026computational,wang2023deviance}. Despite rapid progress in multimodal learning and pathology VLMs, most prior studies treat downstream metrics as primary endpoints. Consequently, feature-level analyses of cross-modal representations in fine-grained renal pathology remain limited, including (i) explicit alignment diagnostics, (ii) class-wise similarity gaps/margins, and (iii) joint visualization of image and text embeddings.

To address these limitations, we explicitly evaluate multimodal representation quality alongside downstream classification performance. In our work, we adapt several general and pathology-specialized VLM backbones to fine-grained glomerular subtyping in a clinically realistic few-shot regime and track how different adaptation strategies reshape the joint image-text embedding space. Beyond reporting Accuracy, AUC, and F1, we quantify paired image-text alignment, class-wise similarity separation, and embedding geometry through visualization and clustering statistics, and we examine how these indicators co-vary with model confidence calibration and cross-center generalization. This feature-level perspective provides a more interpretable basis for selecting VLM backbones and fine-tuning strategies under limited supervision and domain shift.

\subsection{Pathology Benchmarking and Challenge}
In digital pathology, community-driven grand challenges and benchmarks have been key drivers of methodological progress by providing standardized datasets and evaluation protocols that enable objective, like-for-like comparison and improve reproducibility. For example, CAMELYON16/17 established a standardized setting for WSI lymph-node metastasis detection and classification, enabling systematic benchmarking of WSI-level tumor recognition methods~\cite{ehteshami2017diagnostic, bandi2018detection}. More recently, Glo-VLMs extended benchmark-style evaluation to fine-grained renal pathology classification under few-shot regimes, offering a structured view of current VLM capabilities and their limits~\cite{guo2025glo,sohail2025convlm}. Importantly, Glo-VLMs also highlighted remaining evidence gaps and called for stronger feature-level diagnostic analyses, repeated runs to assess stability, and evaluation on external datasets to probe cross-center generalization. These observations align with recent surveys emphasizing the importance of feature analysis, few-shot learning, fine-grained recognition, and cross-center robustness in medical imaging~\cite{xiong2025survey, hosseini2024computational, nazir2023survey, yoon2024domain}. However, existing evaluations typically study these dimensions in isolation across different datasets and experimental settings, lacking a unified and reproducible benchmarking protocol that jointly characterizes these key variables. This fragmentation makes method selection difficult, undermines the stability and comparability of reported gains, and ultimately weakens evidence for clinical usability. 

To address this gap, we introduce a unified benchmark for fine-grained renal pathology classification: under controlled few-shot settings and cross-center external validation, we systematically compare mainstream VLMs and multiple adaptation strategies, and complement downstream metrics with feature-level diagnostic analyses to characterize the performance, representation and generalization relationship, providing a reproducible basis for method comparison and future translational evaluation.

\begin{figure*}[!t]
\centering
\includegraphics[width=0.99\textwidth]{{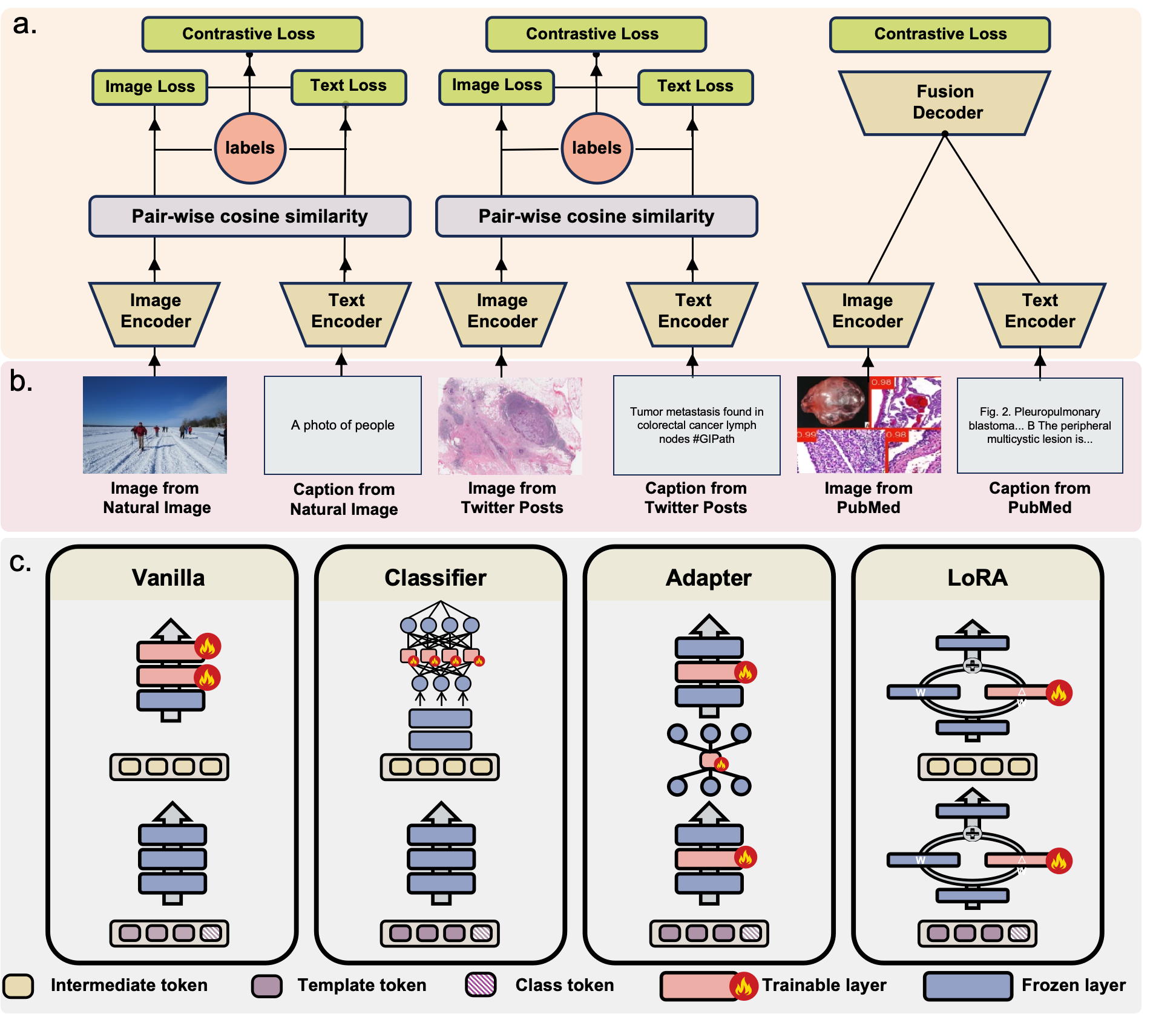}}
\caption{\textbf{(a) Model architectures.} We compare contrastive VLMs with separate image and text encoders trained via pairwise cosine similarity, and a CoCa-style architecture with an additional fusion/decoder branch. \textbf{(b) Domain-specific training data.} We illustrate three sources of paired image-text data: natural images with generic captions, social media posts from Twitter with informal expert commentary, and biomedical figures from PubMed with pathology-style captions. \textbf{(c) Fine-tuning strategies.} We compare (i) Vanilla fine-tuning of the top layers, (ii) LoRA low-rank updates, (iii) trainable adapters between frozen layers, and (iv) a classifier head on a frozen backbone. VLM adaptation uses only the image--text contrastive loss, whereas Classifier tuning uses cross-entropy; no captioning loss is used.}
\label{fig2:Finetuning method}
\end{figure*}
\section{Methods}
	\subsection{Pipeline Overview}
To overcome the limited generalizability of existing methods for fine-grained pathological image classification, our approach centers on distinguishing among subtypes of diseased glomeruli within a low-data setting. Specifically, we investigate the adaptation of VLMs to a five-class glomerular pathology classification task, relying on only a small number of labeled examples per category. As depicted in Fig.~\ref{fig1:Pipeline Overview}, the pipeline begins with pathological image patches paired with their corresponding class prompts (for example, ``A histopathology image of Atubular Glomerulus''). Both the images and textual prompts are encoded using a pretrained VLM backbone, which is adapted to the task either through fine-tuning or the integration of lightweight modules. To reflect the low-data scenario, this adaptation is performed under few-shot conditions, where only a limited set of labeled exemplars per class is available to guide the alignment between modalities. This process yields high-dimensional embeddings for both images and prompts. Except for Classifier tuning, all methods use image--text cosine similarity for inference: we compute the cosine similarity between each image embedding and all class prompt embeddings, applying a softmax transformation to produce a probability distribution over the classes. The model then assigns each image to the class with the highest similarity score. For Classifier tuning, predictions are instead produced by a learned classification head on frozen image features.


\subsection{Foundation VLM Backbones}
To systematically assess the influence of model design, domain expertise, and modality alignment, we select three VLM backbones, as illustrated in Fig.~\ref{fig2:Finetuning method} parts a and b, that differ in scale, architecture, and the extent of pathology-specific training. These are: a general-purpose foundation model (CLIP), a pathology-adapted CLIP model (PLIP), and a pathology-native foundation model (CONCH).
\begin{itemize}
\item \textbf{CLIP}: A large-scale VLM~\cite{radford2021learning}, trained on roughly 400M images with natural language captions using a symmetric contrastive learning objective. In this work, we adopt the openai/clip-vit-base-patch16 variant, which employs a ViT-B/16 image encoder and a Transformer-based text encoder~\cite{dosovitskiy2020image}. Both encoders project their respective inputs into a shared embedding space, normalized for cosine similarity. CLIP enables zero-shot classification and retrieval by comparing the embeddings of images against those of natural language prompts for similarity-based matching.

\item \textbf{PLIP}: A pathology-adapted VLM~\cite{huang2023visual}, developed by retraining CLIP on over 200K pathology image-caption pairs collected primarily from Twitter posts. The model retains the dual-encoder architecture of CLIP, with a ViT-B/16 image encoder and a Transformer-based text encoder, and projects both modalities into a shared embedding space. By aligning histopathology images with domain-specific textual descriptions, PLIP produces embeddings that are more sensitive to pathology-specific features and supports zero-shot classification and cross-modal retrieval in the medical imaging domain.

\item \textbf{CONCH}: A pathology-native VLM~\cite{lu2024visual}, developed on the CoCa architecture~\cite{yu2022coca} with a ViT-B/16 backbone. It is pretrained on 1.17M pathology image-caption pairs curated from PubMed and related pathology resources, combining contrastive alignment and caption generation objectives. By integrating structured attention pooling and a fusion decoder within the CoCa framework, CONCH learns fine-grained multimodal representations of high-resolution histology and supports zero-shot classification and cross-modal retrieval in the pathology domain. In our experiments, we do not use a captioning loss; we fine-tune CONCH using only image--text contrastive learning with fixed class prompts.
\end{itemize}
\begin{figure*}[!t]
\centering
\includegraphics[width=0.7\textwidth]{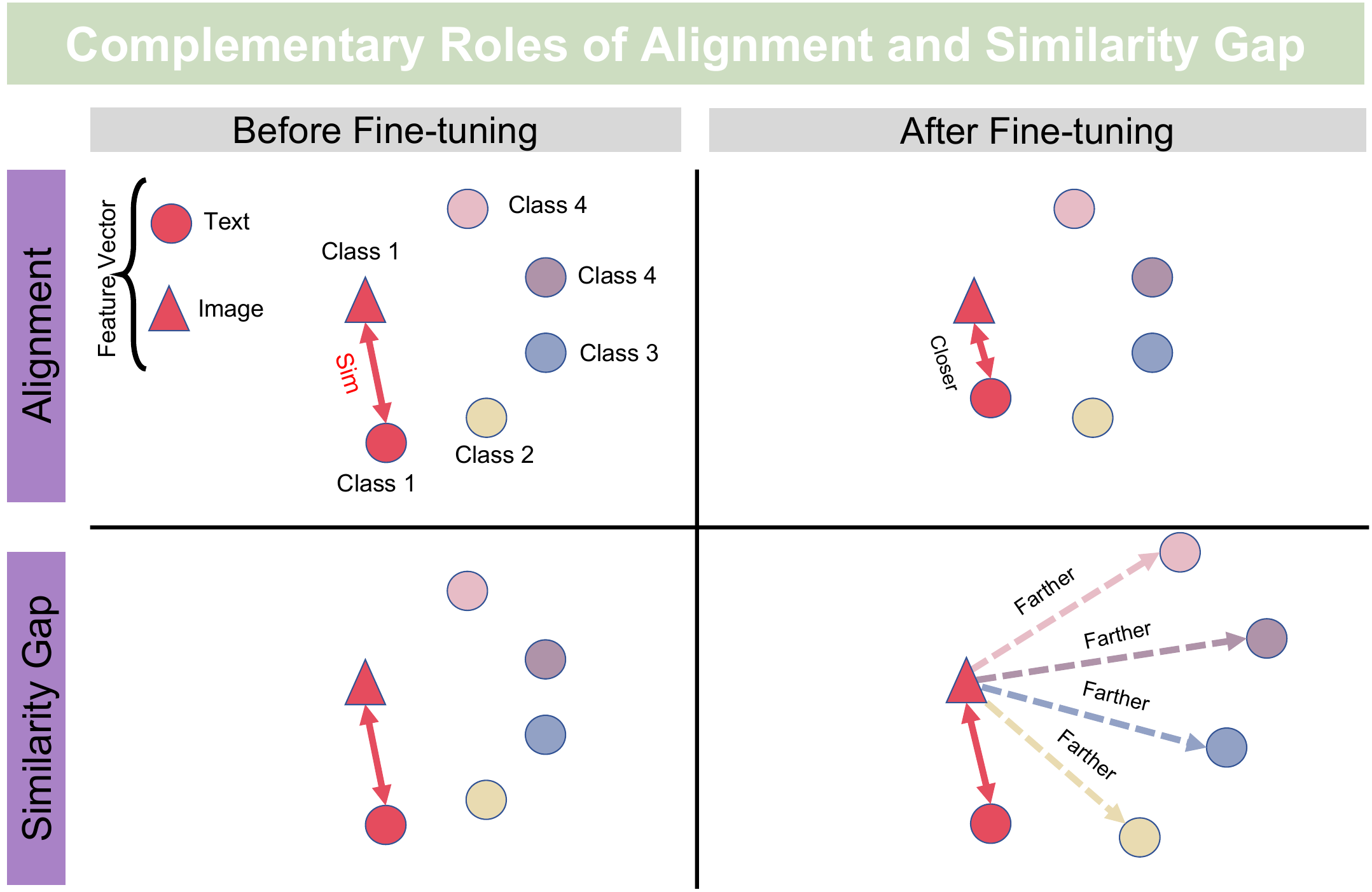}
\caption{\textbf{Complementary roles of alignment and similarity gap.} It shows the change in image and text feature vectors before and after fine-tuning. Alignment measures whether fine-tuning pulls an image embedding closer to the embedding of its own positive text description. Similarity gap is defined as mean positive class-prompt similarity (Alignment) minus mean negative class-prompt similarity, and therefore increases when matched pairs become closer and/or unmatched class prompts become less similar. Together, these two effects capture how fine-tuning reduces within-class image-text distance while enlarging across-class distance.}
\label{fig:align_sim}
\end{figure*}

\begin{figure*}[!t]
\centering
\includegraphics[width=\linewidth]{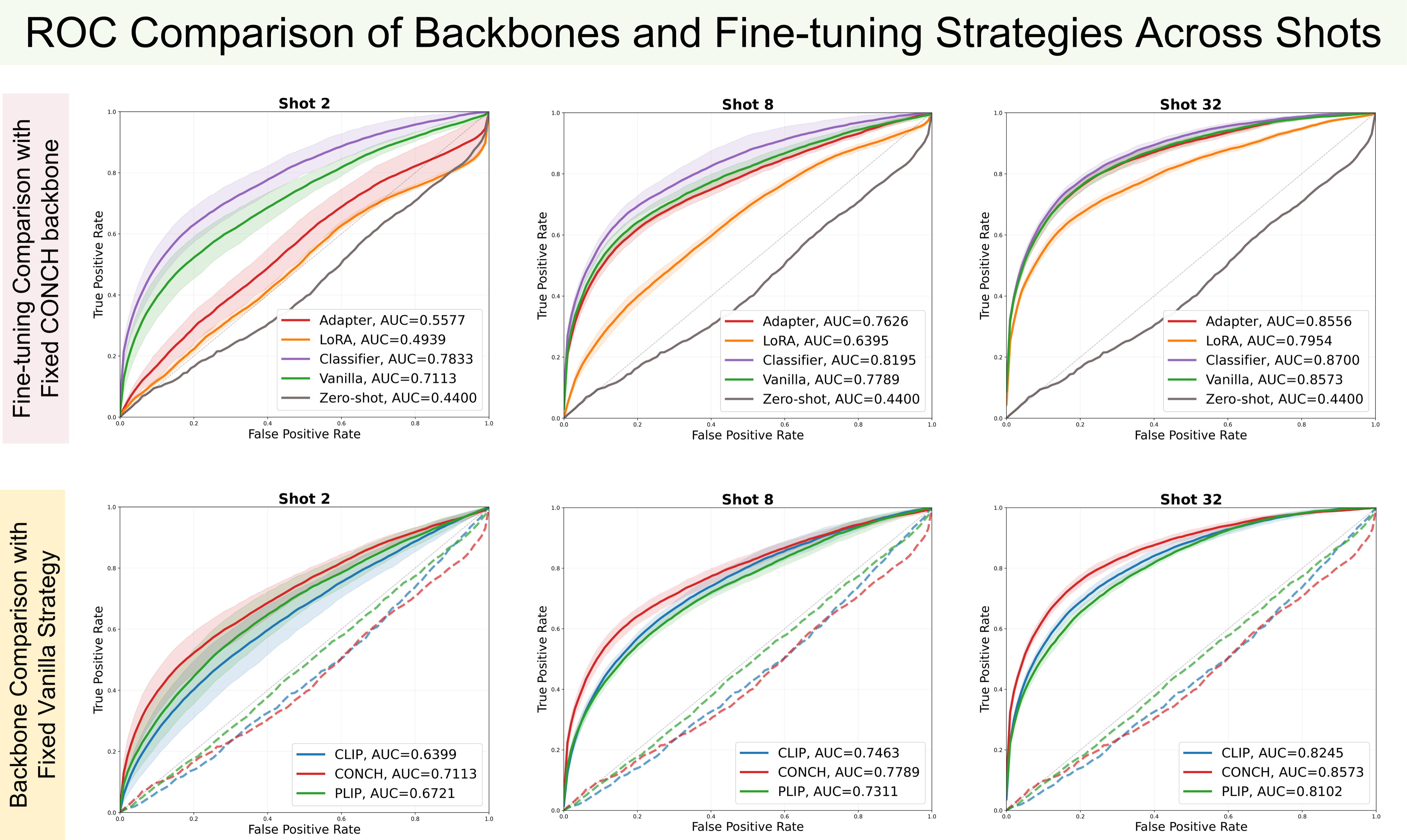}
\caption{\textbf{Receiver operating characteristic (ROC) curves across models and fine-tuning methods on Cornell dataset.}
We compare two axes, model comparison (fixed strategy) and fine-tuning method comparison (fixed backbone), by plotting true-positive rate (TPR) versus false-positive rate (FPR) across increasing shot counts.}
\label{fig:ROC_AUC}
\end{figure*}
\subsection{Fine-tuning Strategies}

To adapt VLMs for fine-grained glomerular classification in few-shot settings, we examine four fine-tuning approaches, as illustrated in Fig.~\ref{fig2:Finetuning method} part c, spanning full-parameter updates to highly parameter-efficient methods~\cite{gao2024clip,srinivasan2024comparative,hu2022lora,ioffe2015batch}. Vanilla fine-tuning modifies both image and text encoders, while Adapter and LoRA insert lightweight modules into the encoder to adjust internal representations with minimal disruption. Classifier tuning updates only the prediction head, keeping the backbone fixed. For VLM adaptation methods (Vanilla, Adapter, and LoRA), we compute class probabilities via cosine similarity between image and text embeddings followed by softmax. Since our task is classification with fixed prompts (not caption generation), we optimize only the image--text contrastive loss and do not use any captioning loss. For Vanilla, Adapter, and LoRA, we use the standard symmetric CLIP image--text contrastive objective~\cite{radford2021learning}. Let $\mathbf{S}\in\mathbb{R}^{N\times N}$ denote the matrix of temperature-scaled cosine similarities for a batch of $N$ paired image and text embeddings, and let $\mathbf{y}^{\mathrm{pair}}=(1,\ldots,N)$ denote the diagonal pairing targets. The final objective is

\begin{equation}
\mathcal{L}_{\mathrm{ITC}}
= \frac{1}{2}\left[
\operatorname{CE}(\mathbf{S},\mathbf{y}^{\mathrm{pair}})
+\operatorname{CE}(\mathbf{S}^{\top},\mathbf{y}^{\mathrm{pair}})
\right].
\label{eq:itc_loss}
\end{equation}
The normalized embeddings, similarity scores, and directional loss terms are detailed in Appendix~\ref{app:metric_definitions}. Classifier tuning uses standard multiclass cross-entropy; its complete definition is also provided in Appendix~\ref{app:metric_definitions}.

\begin{itemize}
\item \textbf{Vanilla Fine-tuning~\cite{srinivasan2024comparative}:}  
This method unfreezes the top layers of the image and text encoders and jointly updates their parameters along with the task-specific head. By training these layers end-to-end, the model adjusts its high-level representations directly to the target task. However, this flexibility comes at the cost of higher computational demands.

\item \textbf{Low-Rank Adaptation (LoRA)~\cite{hu2022lora}:}  
This method injects additional trainable low-rank decomposition matrices into the attention projection layers of the top transformer blocks in both the vision and text encoders. During training, only low-rank decomposition matrices are updated while the original backbone weights remain frozen. The rank $r$ controls the dimensionality of the adaptation subspace, and the scaling factor $\alpha$ regulates the contribution of $\Delta W$ to the final projection. This design reduces the number of trainable parameters while preserving the pretrained backbone.

\item \textbf{Adapter Tuning~\cite{gao2024clip}:}  
This approach preserves all pretrained encoder weights and introduces lightweight bottleneck adapter modules into the Transformer layers~\cite{pfeiffer2020mad}. In our setup, we adopt \emph{sequential bottleneck adapter}, which places a single adapter block only after the feed-forward network of each Transformer layer. Each adapter consists of a down-projection from the hidden dimension $d$ to a smaller bottleneck dimension $m$, a non-linear activation, and an up-projection back to $d$. The transformed output is then added to the residual stream. During training, only the adapter parameters are updated while the backbone remains frozen, enabling parameter-efficient adaptation with minimal additional overhead.

\item \textbf{Classifier Tuning~\cite{ioffe2015batch}:}  
This method freezes all backbone parameters and trains only a lightweight classification head on top of the extracted image features. We evaluate three variants: (i) a single linear projection from the feature dimension to the number of classes; (ii) a two-layer multilayer perceptron (MLP) with a hidden layer and non-linear activation~\cite{guo2025metabolitechat}; and (iii) the same two-layer MLP augmented with batch normalization~\cite{ioffe2015batch} applied after the hidden layer. In all cases, the classifier outputs class logits which are passed through a softmax for prediction. For all reported Classifier tuning results, the final classifier head maps the 512-dimensional frozen image feature to a 512-dimensional hidden layer, followed sequentially by batch normalization, a ReLU activation, and dropout, before a final linear layer produces the $C$ class logits. In our experiments, variant (iii) performs best, and we use it for all reported results.
\end{itemize}

We additionally include two image-only linear-probe baselines, ImageNet-pretrained ResNet-18~\cite{he2016deep} and histopathology-specific self-supervised learning (SSL) ResNet-18~\cite{ciga2022self}, to contextualize the benefits of cross-modal pretraining.

\begin{table*}[!t]
\centering

\begin{adjustbox}{max width=\textwidth}
\begin{tabular}{llccccccc}
\toprule
\multirow{2}{0.4in}{Model} & Method & \multicolumn{7}{c}{Accuracy} \\[0.3em]
\cmidrule{3-9}
& Shots & 0 & 1 & 2 & 4 & 8 & 16 & 32 \\
\midrule

\multirow{4}{*}{CLIP~\cite{radford2021learning}}
&Adapter~\cite{gao2024clip} & 0.3352$\pm$0.0000 & 0.3873$\pm$0.1108 & 0.2960$\pm$0.0881 & 0.4290$\pm$0.1478 & 0.5124$\pm$0.0656 & 0.5617$\pm$0.0385 & 0.5947$\pm$0.0689 \\
&Classifier~\cite{ioffe2015batch} & 0.3352$\pm$0.0000 & 0.1200$\pm$0.0910 & 0.2838$\pm$0.1684 & 0.4445$\pm$0.1244 & 0.4997$\pm$0.1906 & 0.6686$\pm$0.0722 & 0.7140$\pm$0.0317 \\
&LoRA~\cite{hu2022lora} & 0.3352$\pm$0.0000 & 0.3671$\pm$0.1132 & 0.4430$\pm$0.1116 & 0.4930$\pm$0.0829 & 0.5845$\pm$0.0668 & 0.6505$\pm$0.0462 & 0.7028$\pm$0.0655 \\
&Vanilla~\cite{srinivasan2024comparative} & 0.3352$\pm$0.0000 & 0.3505$\pm$0.1088 & 0.4584$\pm$0.1421 & 0.5344$\pm$0.1184 & 0.6441$\pm$0.0728 & 0.6634$\pm$0.0709 & 0.7192$\pm$0.0392 \\
\midrule

\multirow{4}{*}{CONCH~\cite{lu2024visual}}
&Adapter~\cite{gao2024clip} & 0.0363$\pm$0.0000 & 0.0403$\pm$0.0022 & 0.0929$\pm$0.0536 & 0.4446$\pm$0.2094 & 0.6617$\pm$0.0381 & 0.7437$\pm$0.0215 & 0.8082$\pm$0.0224 \\
&Classifier~\cite{ioffe2015batch} & 0.0363$\pm$0.0000 & 0.3468$\pm$0.1912 & 0.4802$\pm$0.1855 & 0.5584$\pm$0.2574 & 0.4664$\pm$0.2487 & 0.7222$\pm$0.0670 & 0.7892$\pm$0.0368 \\
&LoRA~\cite{hu2022lora} & 0.0363$\pm$0.0000 & 0.0379$\pm$0.0009 & 0.0390$\pm$0.0010 & 0.0486$\pm$0.0028 & 0.2806$\pm$0.0567 & 0.6002$\pm$0.0452 & 0.7251$\pm$0.0198 \\
&Vanilla~\cite{srinivasan2024comparative} & 0.0363$\pm$0.0000 & 0.3045$\pm$0.1044 & 0.4764$\pm$0.1280 & \textbf{0.6060}$\pm$0.1045 & \textbf{0.6929}$\pm$0.0409 & \textbf{0.7707}$\pm$0.0400 & \textbf{0.8098}$\pm$0.0160 \\
\midrule

\multirow{4}{*}{PLIP~\cite{huang2023visual}}
&Adapter~\cite{gao2024clip} & 0.3283$\pm$0.0000 & 0.3615$\pm$0.0669 & 0.4200$\pm$0.1471 & 0.5659$\pm$0.0702 & 0.5707$\pm$0.0544 & 0.6034$\pm$0.0577 & 0.6361$\pm$0.0670 \\
&Classifier~\cite{ioffe2015batch} & 0.3283$\pm$0.0000 & \textbf{0.4357}$\pm$0.1658 & 0.4482$\pm$0.1909 & 0.5254$\pm$0.1531 & 0.6170$\pm$0.1088 & 0.6599$\pm$0.0693 & 0.6603$\pm$0.0683 \\
&LoRA~\cite{hu2022lora} & 0.3283$\pm$0.0000 & 0.1894$\pm$0.0739 & 0.2788$\pm$0.0874 & 0.2807$\pm$0.0737 & 0.4696$\pm$0.0876 & 0.5528$\pm$0.0841 & 0.5899$\pm$0.0658 \\
&Vanilla~\cite{srinivasan2024comparative} & 0.3283$\pm$0.0000 & 0.3038$\pm$0.1518 & \textbf{0.4901}$\pm$0.1743 & 0.5830$\pm$0.1387 & 0.5719$\pm$0.1387 & 0.6184$\pm$0.1013 & 0.6256$\pm$0.0622 \\
\midrule

\multirow{1}{*}{ResNet-18~\cite{he2016deep}}
&Supervised Learning & 0.1047$\pm$0.0000 & 0.1078$\pm$0.0016 & 0.1345$\pm$0.0014 & 0.3288$\pm$0.0834 & 0.3466$\pm$0.1019 & 0.2693$\pm$0.0868 & 0.1159$\pm$0.0448 \\
\midrule

\multirow{1}{*}{SSL~\cite{ciga2022self}}
&Supervised Learning & \textbf{0.3491}$\pm$0.0000 & 0.3479$\pm$0.0003 & 0.1523$\pm$0.1667 & 0.2888$\pm$0.1584 & 0.2899$\pm$0.1851 & 0.2968$\pm$0.0874 & 0.1668$\pm$0.1116 \\

\bottomrule
\end{tabular}
\end{adjustbox}

\vspace{1em}

\begin{adjustbox}{max width=\textwidth}
\begin{tabular}{llccccccc}
\toprule
\multirow{2}{0.4in}{Model} & Method & \multicolumn{7}{c}{AUC} \\[0.3em]
\cmidrule{3-9}
& Shots & 0 & 1 & 2 & 4 & 8 & 16 & 32 \\
\midrule

\multirow{4}{*}{CLIP~\cite{radford2021learning}}
&Adapter~\cite{gao2024clip} & 0.4462$\pm$0.0000 & 0.5518$\pm$0.0599 & 0.5766$\pm$0.0478 & 0.6145$\pm$0.0271 & 0.6645$\pm$0.0133 & 0.7075$\pm$0.0324 & 0.7317$\pm$0.0414 \\
&Classifier~\cite{ioffe2015batch} & 0.4462$\pm$0.0000 & 0.6344$\pm$0.0487 & 0.6699$\pm$0.0309 & 0.7171$\pm$0.0273 & 0.7452$\pm$0.0291 & 0.7867$\pm$0.0159 & 0.8064$\pm$0.0153 \\
&LoRA~\cite{hu2022lora} & 0.4462$\pm$0.0000 & 0.5476$\pm$0.0715 & 0.5969$\pm$0.0434 & 0.6284$\pm$0.0274 & 0.7190$\pm$0.0176 & 0.7849$\pm$0.0159 & 0.8148$\pm$0.0117 \\
&Vanilla~\cite{srinivasan2024comparative} & 0.4462$\pm$0.0000 & 0.6132$\pm$0.0534 & 0.6399$\pm$0.0553 & 0.6824$\pm$0.0308 & 0.7463$\pm$0.0161 & 0.7909$\pm$0.0189 & 0.8245$\pm$0.0135 \\
\midrule

\multirow{4}{*}{CONCH~\cite{lu2024visual}}
&Adapter~\cite{gao2024clip} & 0.4400$\pm$0.0000 & 0.4908$\pm$0.0066 & 0.5577$\pm$0.0515 & 0.6727$\pm$0.0911 & 0.7626$\pm$0.0143 & 0.8078$\pm$0.0100 & 0.8556$\pm$0.0110 \\
&Classifier~\cite{ioffe2015batch} & 0.4400$\pm$0.0000 & \textbf{0.7525}$\pm$0.0390 & \textbf{0.7833}$\pm$0.0325 & \textbf{0.8016}$\pm$0.0434 & \textbf{0.8195}$\pm$0.0281 & \textbf{0.8502}$\pm$0.0114 & \textbf{0.8700}$\pm$0.0093 \\
&LoRA~\cite{hu2022lora} & 0.4400$\pm$0.0000 & 0.4877$\pm$0.0039 & 0.4939$\pm$0.0097 & 0.5028$\pm$0.0128 & 0.6395$\pm$0.0142 & 0.7381$\pm$0.0135 & 0.7954$\pm$0.0099 \\
&Vanilla~\cite{srinivasan2024comparative} & 0.4400$\pm$0.0000 & 0.6424$\pm$0.0476 & 0.7113$\pm$0.0331 & 0.7737$\pm$0.0239 & 0.7789$\pm$0.0182 & 0.8294$\pm$0.0269 & 0.8573$\pm$0.0113 \\
\midrule

\multirow{4}{*}{PLIP~\cite{huang2023visual}}
&Adapter~\cite{gao2024clip} & \textbf{0.4825}$\pm$0.0000 & 0.6349$\pm$0.0208 & 0.6120$\pm$0.0731 & 0.6588$\pm$0.0196 & 0.7060$\pm$0.0186 & 0.7377$\pm$0.0234 & 0.7685$\pm$0.0068 \\
&Classifier~\cite{ioffe2015batch} & \textbf{0.4825}$\pm$0.0000 & 0.6719$\pm$0.0496 & 0.6775$\pm$0.0368 & 0.7239$\pm$0.0332 & 0.7419$\pm$0.0230 & 0.7752$\pm$0.0203 & 0.7796$\pm$0.0146 \\
&LoRA~\cite{hu2022lora} & \textbf{0.4825}$\pm$0.0000 & 0.6171$\pm$0.0251 & 0.6298$\pm$0.0454 & 0.6674$\pm$0.0138 & 0.7109$\pm$0.0168 & 0.7755$\pm$0.0163 & 0.8198$\pm$0.0087 \\
&Vanilla~\cite{srinivasan2024comparative} & \textbf{0.4825}$\pm$0.0000 & 0.6425$\pm$0.0359 & 0.6721$\pm$0.0512 & 0.7055$\pm$0.0256 & 0.7311$\pm$0.0176 & 0.7653$\pm$0.0170 & 0.8102$\pm$0.0089 \\
\midrule

\multirow{1}{*}{ResNet-18~\cite{he2016deep}}
&Supervised Learning & 0.4541$\pm$0.0000 & 0.4544$\pm$0.0003 & 0.5437$\pm$0.0004 & 0.5007$\pm$0.0356 & 0.5092$\pm$0.0349 & 0.5014$\pm$0.0569 & 0.4985$\pm$0.0484 \\
\midrule

\multirow{1}{*}{SSL~\cite{ciga2022self}}
&Supervised Learning & 0.4657$\pm$0.0000 & 0.4645$\pm$0.0008 & 0.5832$\pm$0.0895 & 0.4632$\pm$0.0475 & 0.5170$\pm$0.0850 & 0.5112$\pm$0.0729 & 0.5465$\pm$0.0472 \\

\bottomrule
\end{tabular}
\end{adjustbox}

\vspace{1em}

\begin{adjustbox}{max width=\textwidth}
\begin{tabular}{llccccccc}
\toprule
\multirow{2}{0.4in}{Model} & Method & \multicolumn{7}{c}{F1} \\[0.3em]
\cmidrule{3-9}
& Shots & 0 & 1 & 2 & 4 & 8 & 16 & 32 \\
\midrule

\multirow{4}{*}{CLIP~\cite{radford2021learning}}
&Adapter~\cite{gao2024clip} & 0.1080$\pm$0.0000 & 0.1902$\pm$0.0595 & 0.1743$\pm$0.0497 & 0.2425$\pm$0.0692 & 0.2992$\pm$0.0238 & 0.3347$\pm$0.0153 & 0.3495$\pm$0.0356 \\
&Classifier~\cite{ioffe2015batch} & 0.1080$\pm$0.0000 & 0.0722$\pm$0.0489 & 0.1797$\pm$0.0930 & 0.2729$\pm$0.0613 & 0.3080$\pm$0.1029 & 0.3991$\pm$0.0339 & 0.4263$\pm$0.0145 \\
&LoRA~\cite{hu2022lora} & 0.1080$\pm$0.0000 & 0.1738$\pm$0.0511 & 0.2111$\pm$0.0526 & 0.2627$\pm$0.0494 & 0.3505$\pm$0.0187 & 0.4013$\pm$0.0118 & 0.4208$\pm$0.0313 \\
&Vanilla~\cite{srinivasan2024comparative} & 0.1080$\pm$0.0000 & 0.2108$\pm$0.0464 & 0.2480$\pm$0.0695 & 0.3068$\pm$0.0424 & 0.3738$\pm$0.0225 & 0.4055$\pm$0.0238 & 0.4328$\pm$0.0200 \\
\midrule

\multirow{4}{*}{CONCH~\cite{lu2024visual}}
&Adapter~\cite{gao2024clip} & 0.0376$\pm$0.0000 & 0.0439$\pm$0.0036 & 0.0856$\pm$0.0404 & 0.2755$\pm$0.1189 & 0.3892$\pm$0.0177 & 0.4403$\pm$0.0120 & 0.4821$\pm$0.0192 \\
&Classifier~\cite{ioffe2015batch} & 0.0376$\pm$0.0000 & 0.2117$\pm$0.1053 & \textbf{0.3124}$\pm$0.1029 & 0.3248$\pm$0.1704 & 0.3088$\pm$0.1373 & 0.4590$\pm$0.0319 & 0.4915$\pm$0.0291 \\
&LoRA~\cite{hu2022lora} & 0.0376$\pm$0.0000 & 0.0409$\pm$0.0013 & 0.0421$\pm$0.0015 & 0.0509$\pm$0.0025 & 0.2050$\pm$0.0250 & 0.3619$\pm$0.0257 & 0.4331$\pm$0.0113 \\
&Vanilla~\cite{srinivasan2024comparative} & 0.0376$\pm$0.0000 & 0.2086$\pm$0.0464 & 0.2981$\pm$0.0595 & \textbf{0.3799}$\pm$0.0387 & \textbf{0.4109}$\pm$0.0275 & \textbf{0.4640}$\pm$0.0283 & \textbf{0.4957}$\pm$0.0197 \\
\midrule

\multirow{4}{*}{PLIP~\cite{huang2023visual}}
&Adapter~\cite{gao2024clip} & \textbf{0.1492}$\pm$0.0000 & 0.2317$\pm$0.0352 & 0.2482$\pm$0.0829 & 0.3224$\pm$0.0318 & 0.3408$\pm$0.0219 & 0.3596$\pm$0.0222 & 0.3852$\pm$0.0260 \\
&Classifier~\cite{ioffe2015batch} & \textbf{0.1492}$\pm$0.0000 & \textbf{0.2628}$\pm$0.0658 & 0.2726$\pm$0.0766 & 0.3251$\pm$0.0644 & 0.3698$\pm$0.0465 & 0.3986$\pm$0.0280 & 0.4057$\pm$0.0218 \\
&LoRA~\cite{hu2022lora} & \textbf{0.1492}$\pm$0.0000 & 0.1409$\pm$0.0364 & 0.1964$\pm$0.0368 & 0.2154$\pm$0.0305 & 0.3161$\pm$0.0311 & 0.3682$\pm$0.0415 & 0.4038$\pm$0.0273 \\
&Vanilla~\cite{srinivasan2024comparative} & \textbf{0.1492}$\pm$0.0000 & 0.2053$\pm$0.0781 & 0.2815$\pm$0.0836 & 0.3423$\pm$0.0539 & 0.3489$\pm$0.0472 & 0.3809$\pm$0.0402 & 0.4161$\pm$0.0233 \\
\midrule

\multirow{1}{*}{ResNet-18~\cite{he2016deep}}
&Supervised Learning & 0.0679$\pm$0.0000 & 0.0693$\pm$0.0007 & 0.0746$\pm$0.0013 & 0.1165$\pm$0.0366 & 0.1576$\pm$0.0370 & 0.1289$\pm$0.0321 & 0.0996$\pm$0.0321 \\
\midrule

\multirow{1}{*}{SSL~\cite{ciga2022self}}
&Supervised Learning & 0.1210$\pm$0.0000 & 0.1221$\pm$0.0001 & 0.0556$\pm$0.0584 & 0.1233$\pm$0.0455 & 0.1564$\pm$0.0844 & 0.1204$\pm$0.0229 & 0.1002$\pm$0.0505 \\

\bottomrule
\end{tabular}
\end{adjustbox}

\caption{\textbf{Few-shot classification performance on Cornell.}
Mean$\pm$SD Accuracy, AUC, and F1 across shot settings $\{0,1,2,4,8,16,32\}$ for all backbone and adaptation pairs.
Here, $k$-shot denotes $k$ labeled images per class used for fine-tuning. For the VLMs, 0-shot denotes inference without few-shot adaptation; for the ResNet-18 linear-probe baselines, 0-shot denotes the reference result obtained with a randomly initialized linear classification head before any few-shot training.
Bold indicates the best value within each shot column (higher is better).}
\label{tab:table_1}
\end{table*}

\begin{figure*}[!t]
\centering
\includegraphics[width=\linewidth]{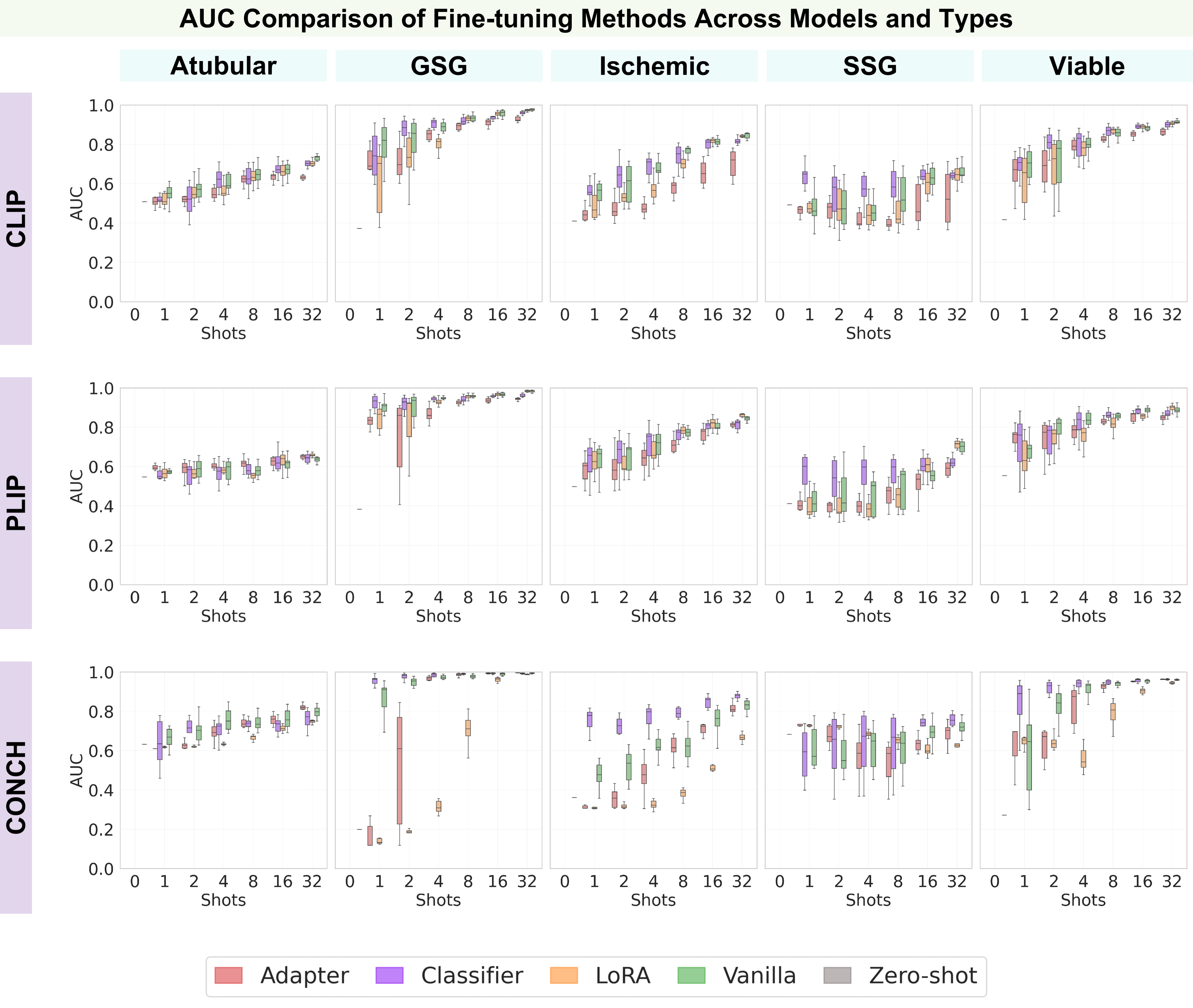}
\caption{\textbf{Discrimination across subtypes on Cornell.} This figure summarizes per-class discrimination performance across 10 runs. For each glomerular subtype, we report the distribution of AUC across VLM backbones, fine-tuning strategies, and shot settings. GSG and SSG denote Globally Sclerotic Glomerulus and Segmentally Sclerotic Glomerulus, respectively.}
\label{fig:boxplot}
\end{figure*}

\begin{figure*}[!t]
\centering
\includegraphics[width=\linewidth]{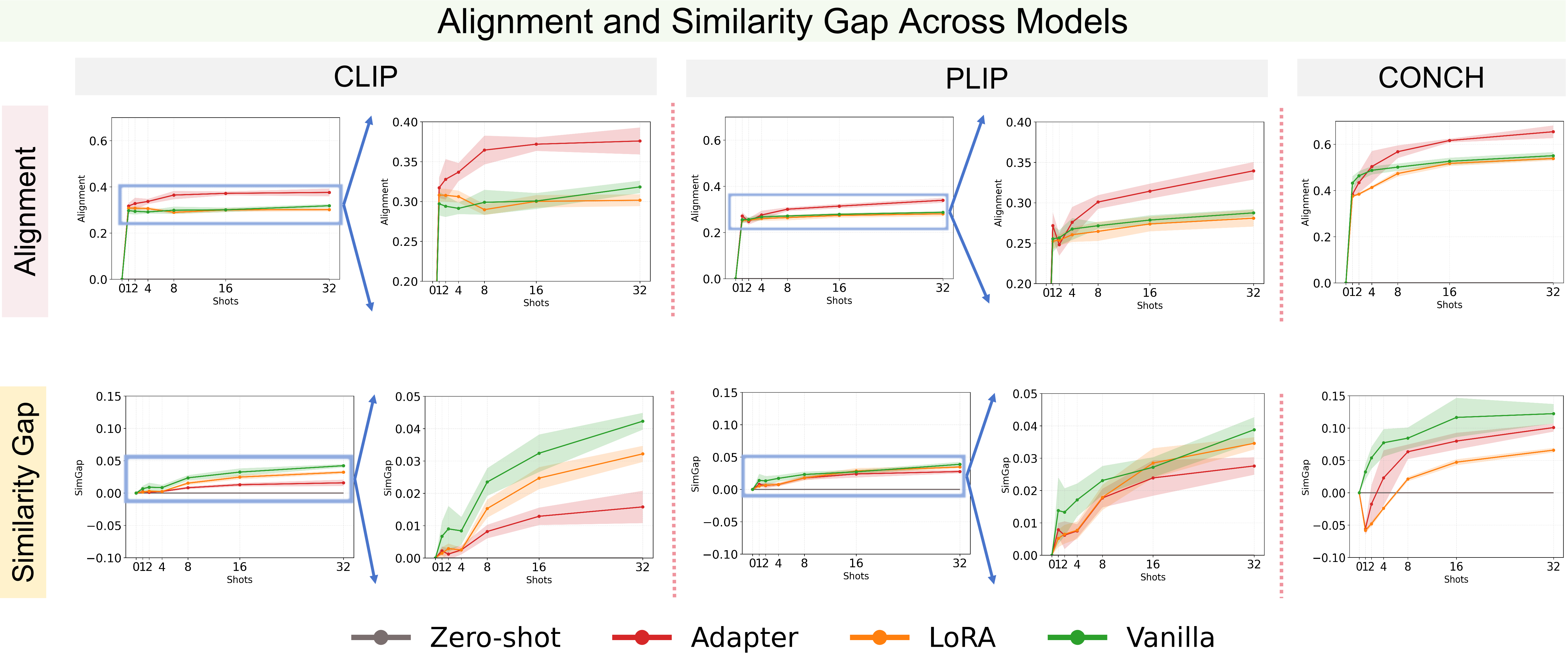}
\caption{\textbf{Alignment and similarity gap across models on Cornell.} This figure shows shot-wise trajectories of Alignment and Similarity Gap for CLIP, PLIP, and CONCH under different fine-tuning strategies. Similarity Gap is defined as mean positive similarity (Alignment) minus mean negative similarity. CLIP and PLIP are demonstrated in both regular and enlarged panels to show their behavior. }
\label{fig:align_sim_result}
\end{figure*}

\section{Experiments and Data}
\subsection{Data and Composition}

We conducted experiments on two renal WSI cohorts, referred to as the Cornell dataset and the Vanderbilt dataset.

\subsubsection{Cornell Dataset}
The Cornell cohort retrospectively includes the non-tumoral kidney parenchyma of eight radical nephrectomy cases, yielding a total of 17 WSIs, processed at the Department of Pathology, Northwell Health, with specimens sourced from Long Island Jewish Medical Center and Northshore University Hospital (July 2017 to October 2019). Using the provided annotation masks, we extracted 6{,}726 glomerulus-centered patches and assigned each patch to one of five glomerular subtypes: Atubular glomeruli (230, 3.42\%), Global glomerulosclerosis (2{,}500, 37.17\%), Ischemic glomeruli (502, 7.46\%), Segmental glomerulosclerosis (271, 4.03\%), and Viable glomeruli (3{,}223, 47.92\%).

\subsubsection{Vanderbilt Dataset}
The Vanderbilt cohort comprises 141 renal WSIs from 126 patients with linked patient-level clinical information. Tissue specimens were processed as formalin-fixed, paraffin-embedded blocks, sectioned at \SI{3}{\micro\meter}, and stained with Periodic Acid-Schiff. The cohort included 87 males and 39 females (mean age 63.4 $\pm$ 12.0 years; median 63.8 years) and was predominantly Caucasian (92.9\%), followed by African American (6.3\%) and Asian-Pacific (0.8\%). Available clinical annotations included serum creatinine, blood urea nitrogen, creatinine clearance, hypertension, and diabetes; mean (median) serum creatinine, blood urea nitrogen, and creatinine clearance were 1.68 (1.02) mg/dL, 22.82 (18.00) mg/dL, and 66.42 (72.74) mL/min, respectively. From the WSIs, we extracted 13{,}922 patches and assigned each patch to one of five categories: Disappearing glomeruli (434, 3.12\%), Non-glomerular (4{,}865, 34.94\%), Normal glomeruli (3{,}609, 25.92\%), Obsolescent glomeruli (4{,}289, 30.81\%), and Solidified glomeruli (725, 5.21\%).

\subsection{Data Preprocessing}
To standardize both the physical sampling resolution and the model input dimensions across cohorts, image patches were generated at a common target resolution of approximately 0.50 $\mu$m/pixel. The Cornell WSIs were acquired at 20$\times$ magnification with a native pixel spacing of approximately 0.50 $\mu$m/pixel, whereas the Vanderbilt WSIs were acquired at 40$\times$ magnification with a native pixel spacing of approximately 0.25 $\mu$m/pixel, as recorded in the WSI metadata. During patch extraction, Vanderbilt image regions were downsampled to approximately 0.50 $\mu$m/pixel to match the Cornell cohort. Individual glomeruli were localized using the provided annotation masks, and an annotation-derived bounding box was computed for each glomerulus. The center of this bounding box was then used as the center of a fixed 512 $\mu$m $\times$ 512 $\mu$m square field of view. Sampling this field at approximately 0.50 $\mu$m/pixel produced a 1024 $\times$ 1024-pixel patch for downstream training. This procedure ensured consistent physical sampling resolution and field of view across both cohorts.

For the textual component, we generate class-specific prompts that serve as anchors for contrastive learning. Each of the five glomerular subtypes is represented by a standardized phrase in the form: ``A histopathology image of \{label\}.'' The text prompt is kept fixed in all experiments for consistency in image-text alignment.

To ensure valid few-shot evaluation, we build class-balanced training sets with 1, 2, 4, 8, 16, and 32 shots per class. We enforce strict WSI-level separation and patient-level isolation. All training patches come only from training WSIs. The validation set uses only WSIs that never appear in training. For both datasets, we used the same WSI-level split strategy: 2 WSIs for training, 2 WSIs for validation, and all remaining WSIs for testing. The validation set therefore contains 136 samples for Vanderbilt and 790 samples for Cornell. The few-shot training set contains 1, 2, 4, 8, 16, or 32 samples per class, corresponding to 5, 10, 20, 40, 80, or 160 total training samples per run, respectively. For each of the 10 independent runs, we constructed a distinct class-balanced training set from the same training WSIs, while keeping the validation and test WSIs fixed. For each run and class, we first form a 32-sample superset from the training WSIs. We fix a per-run permutation and derive smaller shots as prefixes of this order, where each smaller-shot subset is fully contained within all larger-shot sets. This design guarantees that, for example, the 4-shot configuration contains the exact images from the 2-shot set plus two additional images per class, enabling a controlled comparison across shot levels. We also maximize diversity across runs, especially for small shots. Each run receives a distinct round-robin slice of the per-class pool, and we prioritize that slice to reduce overlap across runs. All sampling uses fixed random seeds and deterministic rules.
\subsection{Experimental Design}
To ensure robust and stable model optimization, we performed a systematic exploration of key hyperparameters. Training was conducted using the AdamW optimizer with an initial learning rate of \(3\times10^{-4}\), incorporating a linear warm-up stage followed by scheduled decay to ensure smooth and stable convergence. To further strengthen generalization and mitigate overfitting, we uniformly applied data augmentation using the Albumentations~\cite{info11020125} library, including vertical and horizontal flips, transpose, random \(90^\circ\) rotation, blur-based perturbations (MotionBlur, MedianBlur, Blur), and contrast/texture adjustments (CLAHE, RandomBrightnessContrast, Sharpen), and employed early stopping with consistent patience and minimum-delta settings across all experiments. For parameter-efficient adaptation, we set LoRA rank to \(r=16\) and scaling factor to \(\alpha=32\), and set the adapter bottleneck hidden dimension to \(m=64\). For Classifier tuning, we used an MLP BatchNorm head with one hidden layer; the input feature dimension is \(512\). We optimized only the classifier-head parameters using AdamW with a learning rate of \(1\times10^{-3}\) and weight decay of \(1\times10^{-4}\); the hidden dimension was set to \(512\), and the dropout rate was set to \(0.1\). The ResNet-18 and SSL baselines use the same WSI-level splits and few-shot budgets as the VLM experiments. All experiments were conducted on an NVIDIA RTX 6000 Ada Generation with 48 GB of VRAM. For quantitative results, we report the mean and standard deviation over 10 independent runs. Uniform Manifold Approximation and Projection (UMAP) visualizations are generated from the run achieving the highest AUC.

\subsection{Evaluation Metrics}
We evaluated all methods on a five-class glomerular classification task using a fixed train-validation split and standardized class prompts. Performance was reported using overall accuracy, macro-AUC, and F1 score, with identical data augmentation and early stopping applied to ensure a fair comparison~\cite{wang2024multiclass}. 

In addition to standard metrics, we further assessed internal representation alignment by measuring the alignment and similarity gap, as depicted in Fig.~\ref{fig:align_sim}. Alignment quantifies the average cosine similarity between paired image-text embeddings, reflecting how well the two modalities are semantically matched, and is computed in Eq.~\ref{eq:alignment}:

\begin{equation}
\mathrm{Alignment}=
\frac{1}{N}\sum_{i=1}^{N}
\cos(\mathbf{z}_i,\mathbf{w}_{y_i}).
\label{eq:alignment}
\end{equation}

The similarity gap measures the mean separation between the positive class prompt and the negative class prompts, capturing the model's discriminative ability in distinguishing aligned from unaligned class-level representations and is defined in Eq.~\ref{eq:sim_gap}:

\begin{equation}
\begin{aligned}
\mathrm{Similarity\ Gap}
&=\frac{1}{N}\sum_{i=1}^{N}\cos(\mathbf{z}_i,\mathbf{w}_{y_i})\\
&\quad-\frac{1}{N(C-1)}
\sum_{i=1}^{N}
\sum_{\substack{c=1 \\ c\ne y_i}}^{C}
\cos(\mathbf{z}_i,\mathbf{w}_{c}).
\end{aligned}
\label{eq:sim_gap}
\end{equation}

\noindent
where $\mathbf{z}_i$ denotes the normalized image embedding of the $i$-th sample, $y_i$ is its class label, $C$ is the number of classes, and $\mathbf{w}_{c}$ denotes the normalized text embedding of the prompt for class $c$. The negative term averages over all class prompts other than the ground-truth class prompt for each image. Therefore, the similarity gap increases when positive class-prompt similarity rises, negative class-prompt similarity decreases, or both. Together, these two metrics provide complementary perspectives on multimodal consistency. We also visualized multimodal feature distributions using UMAP for dimensionality reduction and kernel density estimation (KDE) for density-based visualization to provide qualitative insights. UMAP was chosen as it provides a compact yet faithful two-dimensional representation of the embedding space, facilitating visual comparison of inter-class and cross-modal distributions. All UMAP embeddings and KDE visualizations were produced using consistent settings across experiments.

Beyond alignment-based measures, we report Intra-Class Distance and Silhouette Score. For intra-class distance, let $\mathbf{x}_i$ denote the image representation used for geometric analysis and $y_i$ its class label. For Classifier tuning, $\mathbf{x}_i$ is the penultimate hidden representation of the tuned MLP head fed by fixed 512-D CONCH features; for the other fine-tuning strategies, $\mathbf{x}_i$ is the corresponding image embedding produced by the tuned encoder. We obtain the two-dimensional UMAP projection $\mathbf{u}_i=\mathrm{UMAP}(\mathbf{x}_i)$ using fixed hyperparameters and seed across experiments. Intra-class distance is the mean pairwise euclidean distance among same-class samples in the UMAP space, as shown in Eq.~\ref{eq:icd}:

\begin{equation}
\mathrm{ICD}
=\frac{1}{|\mathcal{C}|}\sum_{c\in\mathcal{C}}
\frac{2}{n_c(n_c-1)}
\sum_{\substack{i<j\\ y_i=y_j=c}}
\|\mathbf{u}_i-\mathbf{u}_j\|_2.
\label{eq:icd}
\end{equation}

\noindent
where $\mathcal{C}$ is the set of classes and $n_c$ is the number of samples in class $c$.
To assess separability in the same representation space, we compute the Silhouette score~\cite{rousseeuw1987silhouettes} on $\{\mathbf{x}_i\}$ with cosine distance. For each sample, $a_i$ denotes its mean distance to samples in the same class, whereas $b_i$ denotes its minimum mean distance to any other class. The final score is

\begin{equation}
S=\frac{1}{N}\sum_{i=1}^{N}
\frac{b_i-a_i}{\max\{a_i,b_i\}},
\qquad S\in[-1,1].
\label{eq:silhouette}
\end{equation}
Detailed definitions of $a_i$ and $b_i$ are provided in Appendix~\ref{app:metric_definitions}.

To quantify model confidence and better characterize decision boundaries, we additionally evaluate ECE~\cite{guo2017calibration} and boundary margin, defined as the top-1 vs. top-2 probability gap. We compute top-1 ECE using $M=15$ equal-width confidence bins over $[0,1]$:
\begin{equation}
\mathrm{ECE}=\sum_{m=1}^{M}\frac{|B_m|}{N}\,
\big|\mathrm{acc}(B_m)-\mathrm{conf}(B_m)\big|.
\label{eq:ece}
\end{equation}
Here, $N$ is the number of test samples and $|B_m|$ is the number of samples in bin $B_m$; lower ECE indicates better-calibrated predicted probabilities. The bin-level accuracy and confidence definitions are provided in Appendix~\ref{app:metric_definitions}.

We quantify boundary margin as a simple proxy for decision boundary sharpness in the model's predicted probability space. For each sample $i$ with predicted class probabilities $\mathbf{p}_i=\mathrm{softmax}(\mathbf{s}_i)$, let $p_i^{(1)}$ and $p_i^{(2)}$ denote the largest and second-largest entries in $\mathbf{p}_i$, respectively. The boundary margin is defined in Eq.~\ref{eq:prob_gap}:
\begin{equation}
\Delta p_i = p_i^{(1)} - p_i^{(2)}.
\label{eq:prob_gap}
\end{equation}
Larger $\Delta p_i$ indicates a clearer separation between the top prediction and the closest alternative, suggesting that the sample lies farther from the decision boundary in the model's predicted probability space.

For clinical label-quality assessment, we additionally performed an expert label audit on a balanced subset of 500 Cornell patches, with 100 patches sampled from each of the five classes. Three doctors independently reviewed the class assignment of each patch using the same five-class schema. We quantified inter-rater reliability using Fleiss' $\kappa$~\cite{fleiss1971measuring}, which measures agreement beyond chance. Its final form is
\begin{equation}
\kappa=\frac{\bar{P}-\bar{P}_e}{1-\bar{P}_e},
\label{eq:fleiss_kappa}
\end{equation}
where $\bar{P}$ is the mean observed agreement and $\bar{P}_e$ is the expected agreement by chance. Their detailed computation is provided in Appendix~\ref{app:metric_definitions}. In our study, $N=500$, $n=3$, and $k=5$. We computed physician recall within each ground-truth class and a one-vs-rest Fleiss' $\kappa$ for each class by collapsing that class against the remaining four classes. The Average entry in the class-wise $\kappa$ row is therefore a macro-average across the five one-vs-rest Fleiss' $\kappa$ values.

\begin{figure*}[!t]
\centering
\includegraphics[width=\linewidth]{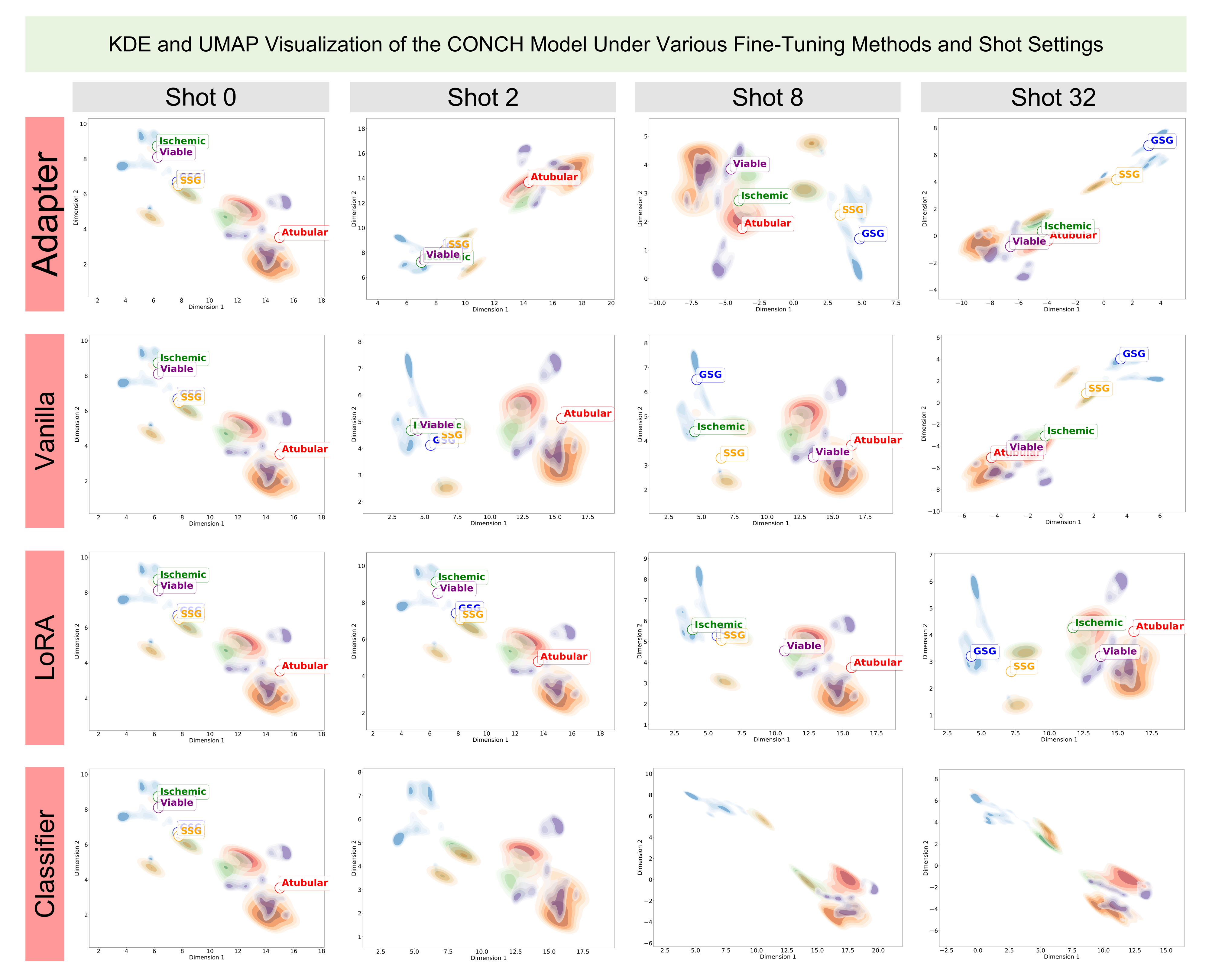}
\caption{\textbf{UMAP/KDE visualization of embeddings across shots on Cornell.}
This figure shows text and image embeddings from the best CONCH run across shot settings. For Classifier tuning, text is not updated and only image-side embeddings are shown. Specifically, for Classifier tuning we visualize the penultimate hidden representation of the tuned MLP head (fed by fixed 512-D CONCH features), which changes as the head is updated. The companion Table~\ref{tab:tabel2} reports Intra-Class Distance and Silhouette Score.}
\label{fig:umap}
\end{figure*}

\begin{table*}[!t]
\centering
\begin{adjustbox}{max width=\textwidth}
\begin{tabular}{llcccccccccccccc}
\toprule
\multirow{2}{0.4in}{Model} & Method & \multicolumn{7}{c}{Intra-Class Distance} & \multicolumn{7}{c}{Silhouette Score} \\[0.3em]
\cmidrule{3-9}
\cmidrule{10-16}
& Shots & 0 & 1 & 2 & 4 & 8 & 16 & 32 & 0 & 1 & 2 & 4 & 8 & 16 & 32 \\
\midrule

\multirow{4}{*}{CLIP~\cite{radford2021learning}}
& Adapter~\cite{gao2024clip} & \textbf{2.5430} & 2.4421 & 2.4591 & 2.5170 & 3.1200 & 2.6249 & 2.4688 & 0.0373 & 0.0330 & 0.0318 & 0.0307 & 0.0701 & 0.0677 & 0.0803 \\
& Classifier~\cite{ioffe2015batch} & \textbf{2.5430} & 2.4863 & 2.3897 & 2.6498 & 2.7732 & 2.8545 & 2.9882 & 0.0373 & 0.0490 & 0.0292 & 0.0718 & 0.0884 & 0.0807 & 0.0968 \\
& LoRA~\cite{hu2022lora} & \textbf{2.5430} & \textbf{2.3812} & 2.4981 & 2.5857 & 3.0166 & 2.5515 & 2.4130 & 0.0373 & 0.0332 & 0.0310 & 0.0331 & 0.0591 & 0.0557 & 0.0968 \\
& Vanilla~\cite{srinivasan2024comparative} & \textbf{2.5430} & 2.6376 & \textbf{2.3177} & \textbf{2.3004} & \textbf{2.5726} & \textbf{2.3463} & 2.6607 & 0.0373 & 0.0440 & 0.0279 & 0.0410 & 0.0773 & 0.0665 & 0.1252 \\
\midrule

\multirow{4}{*}{CONCH~\cite{lu2024visual}}
& Adapter~\cite{gao2024clip} & 3.0591 & 3.1978 & 3.4481 & 3.1913 & 3.0749 & 3.0657 & 2.8749 & \textbf{0.1812} & 0.1794 & 0.1729 & 0.1614 & 0.1931 & 0.2150 & 0.3158 \\
& Classifier~\cite{ioffe2015batch} & 3.0591 & 2.9927 & 3.0976 & 2.8299 & 3.6277 & 3.6330 & 3.6131 & \textbf{0.1812} & \textbf{0.1870} & \textbf{0.2002} & 0.1775 & 0.2929 & 0.2301 & 0.2887 \\
& LoRA~\cite{hu2022lora} & 3.0591 & 3.1626 & 3.0679 & 3.1751 & 3.2274 & 3.1471 & 3.0469 & \textbf{0.1812} & 0.1805 & 0.1806 & 0.1800 & 0.1774 & 0.1767 & 0.1873 \\
& Vanilla~\cite{srinivasan2024comparative} & 3.0591 & 3.3166 & 3.2381 & 3.0948 & 3.1859 & 2.8229 & \textbf{2.3254} & \textbf{0.1812} & 0.1864 & 0.1792 & \textbf{0.2041} & \textbf{0.3027} & \textbf{0.3453} & \textbf{0.3923} \\
\midrule

\multirow{4}{*}{PLIP~\cite{huang2023visual}}
& Adapter~\cite{gao2024clip} & 4.6734 & 4.7702 & 5.3012 & 4.5804 & 4.6559 & 4.5383 & 4.3455 & 0.0768 & 0.0801 & 0.0768 & 0.0677 & 0.1049 & 0.0900 & 0.0975 \\
& Classifier~\cite{ioffe2015batch} & 4.6734 & 4.0596 & 5.0829 & 4.5074 & 4.2935 & 4.4788 & 4.4356 & 0.0768 & 0.1336 & 0.1356 & 0.1350 & 0.1314 & 0.1450 & 0.1548 \\
& LoRA~\cite{hu2022lora} & 4.6734 & 4.8622 & 4.3826 & 5.0877 & 4.7062 & 4.7701 & 5.0552 & 0.0768 & 0.0711 & 0.0703 & 0.0687 & 0.0915 & 0.0964 & 0.1260 \\
& Vanilla~\cite{srinivasan2024comparative} & 4.6734 & 4.6071 & 4.6863 & 5.1612 & 4.7397 & 4.8918 & 4.9956 & 0.0768 & 0.0790 & 0.0762 & 0.0862 & 0.0951 & 0.1109 & 0.1214 \\
\midrule

\bottomrule
\end{tabular}
\end{adjustbox}
\caption{\textbf{Intra-Class Distance and Silhouette Score on Cornell.}
Few-shot geometric metrics under three backbones: CLIP, CONCH, and PLIP across shot settings $\{0,1,2,4,8,16,32\}$ and four fine-tuning strategies.
Intra-Class Distance measures within-class compactness in the UMAP space (lower is better), and Silhouette Score measures cluster separability in the original embedding space using cosine distance (higher is better). For Classifier tuning, both metrics are computed from the penultimate representation of the tuned MLP head.}
\label{tab:tabel2}
\end{table*}

\begin{table*}[t]
\centering
\begin{adjustbox}{max width=\textwidth}

\begin{tabular}{llccccccc}
\toprule
\multirow{2}{0.4in}{Model} & Method & \multicolumn{7}{c}{ECE (Top-1)} \\[0.3em]
\cmidrule{3-9}
& Shots & 0 & 1 & 2 & 4 & 8 & 16 & 32 \\
\midrule

\multirow{4}{*}{CLIP~\cite{radford2021learning}}
&Adapter~\cite{gao2024clip} & 0.1220$\pm$0.0000 & 0.1336$\pm$0.0583 & \textbf{0.0832}$\pm$0.0511 & 0.1926$\pm$0.0830 & 0.1784$\pm$0.0450 & 0.1460$\pm$0.0370 & 0.1307$\pm$0.0490 \\
&Classifier~\cite{ioffe2015batch} & 0.1220$\pm$0.0000 & 0.1317$\pm$0.0625 & 0.1067$\pm$0.0370 & 0.1411$\pm$0.0984 & 0.0932$\pm$0.0471 & \textbf{0.0618}$\pm$0.0312 & \textbf{0.0476}$\pm$0.0219 \\
&LoRA~\cite{hu2022lora} & 0.1220$\pm$0.0000 & 0.1119$\pm$0.0523 & 0.1631$\pm$0.0520 & 0.2231$\pm$0.0716 & 0.0798$\pm$0.0271 & 0.0641$\pm$0.0271 & 0.0525$\pm$0.0428 \\
&Vanilla~\cite{srinivasan2024comparative} & 0.1220$\pm$0.0000 & 0.1702$\pm$0.0726 & 0.1150$\pm$0.0848 & 0.1597$\pm$0.0813 & 0.0741$\pm$0.0377 & 0.0822$\pm$0.0489 & 0.0829$\pm$0.0341 \\
\midrule

\multirow{4}{*}{CONCH~\cite{lu2024visual}}
&Adapter~\cite{gao2024clip} & 0.1790$\pm$0.0000 & 0.8659$\pm$0.0064 & 0.7665$\pm$0.0889 & 0.3776$\pm$0.2461 & 0.1841$\pm$0.0353 & 0.1462$\pm$0.0238 & 0.1234$\pm$0.0178 \\
&Classifier~\cite{ioffe2015batch} & 0.1790$\pm$0.0000 & 0.1875$\pm$0.1001 & 0.2588$\pm$0.1676 & 0.3281$\pm$0.1150 & 0.2727$\pm$0.1230 & 0.3610$\pm$0.1562 & 0.2004$\pm$0.1815 \\
&LoRA~\cite{hu2022lora} & 0.1790$\pm$0.0000 & 0.8672$\pm$0.0013 & 0.8657$\pm$0.0021 & 0.8349$\pm$0.0121 & 0.5096$\pm$0.0671 & 0.2203$\pm$0.0330 & 0.1575$\pm$0.0167 \\
&Vanilla~\cite{srinivasan2024comparative} & 0.1790$\pm$0.0000 & 0.4287$\pm$0.1241 & 0.2662$\pm$0.1084 & 0.1938$\pm$0.0695 & 0.1255$\pm$0.0405 & 0.1104$\pm$0.0262 & 0.0954$\pm$0.0192 \\
\midrule

\multirow{4}{*}{PLIP~\cite{huang2023visual}}
&Adapter~\cite{gao2024clip} & 0.1120$\pm$0.0000 & 0.1102$\pm$0.0501 & 0.1022$\pm$0.0888 & 0.1736$\pm$0.0550 & \textbf{0.0658}$\pm$0.0203 & 0.0763$\pm$0.0423 & 0.0582$\pm$0.0383 \\
&Classifier~\cite{ioffe2015batch} & 0.1120$\pm$0.0000 & 0.2462$\pm$0.0928 & 0.2587$\pm$0.1440 & 0.1898$\pm$0.1135 & 0.1523$\pm$0.1122 & 0.0826$\pm$0.0540 & 0.0780$\pm$0.0514 \\
&LoRA~\cite{hu2022lora} & 0.1120$\pm$0.0000 & 0.3033$\pm$0.0983 & 0.1900$\pm$0.0762 & 0.1710$\pm$0.0563 & 0.1203$\pm$0.0584 & 0.1388$\pm$0.0824 & 0.1407$\pm$0.0672 \\
&Vanilla~\cite{srinivasan2024comparative} & 0.1120$\pm$0.0000 & 0.3047$\pm$0.1390 & 0.1433$\pm$0.0988 & 0.1161$\pm$0.0988 & 0.1347$\pm$0.0901 & 0.1023$\pm$0.0794 & 0.1226$\pm$0.0622 \\
\midrule

\multirow{1}{*}{ResNet-18~\cite{he2016deep}}
&Supervised Learning & 0.2614$\pm$0.0000 & 0.2523$\pm$0.0015 & 0.2496$\pm$0.0025 & \textbf{0.1039}$\pm$0.0364 & 0.1005$\pm$0.0359 & 0.0889$\pm$0.0678 & 0.1655$\pm$0.0494 \\
\midrule

\multirow{1}{*}{SSL~\cite{ciga2022self}}
&Supervised Learning & \textbf{0.0877}$\pm$0.0000 & \textbf{0.0896}$\pm$0.0004 & 0.2444$\pm$0.0280 & 0.1550$\pm$0.0519 & 0.1619$\pm$0.1059 & 0.1056$\pm$0.0441 & 0.1104$\pm$0.0765 \\

\bottomrule
\end{tabular}
\end{adjustbox}

\vspace{1em}

\begin{adjustbox}{max width=\textwidth}
\begin{tabular}{llccccccc}
\toprule
\multirow{2}{0.4in}{Model} & Method & \multicolumn{7}{c}{Boundary Margin (Mean)} \\[0.3em]
\cmidrule{3-9}
& Shots & 0 & 1 & 2 & 4 & 8 & 16 & 32 \\
\midrule

\multirow{4}{*}{CLIP~\cite{radford2021learning}}
&Adapter~\cite{gao2024clip} & 0.0006$\pm$0.0000 & 0.0589$\pm$0.0185 & 0.0459$\pm$0.0220 & 0.0525$\pm$0.0197 & 0.1491$\pm$0.0294 & 0.2230$\pm$0.0484 & 0.2762$\pm$0.1015 \\
&Classifier~\cite{ioffe2015batch} & 0.0006$\pm$0.0000 & 0.0181$\pm$0.0156 & 0.1023$\pm$0.1293 & 0.1488$\pm$0.1683 & 0.3085$\pm$0.2280 & 0.5333$\pm$0.1413 & 0.5879$\pm$0.0519 \\
&LoRA~\cite{hu2022lora} & 0.0006$\pm$0.0000 & 0.0642$\pm$0.0247 & 0.0773$\pm$0.0342 & 0.0520$\pm$0.0132 & 0.3226$\pm$0.0685 & 0.4614$\pm$0.0681 & 0.5753$\pm$0.0503 \\
&Vanilla~\cite{srinivasan2024comparative} & 0.0006$\pm$0.0000 & 0.2218$\pm$0.1082 & 0.2319$\pm$0.1293 & 0.1751$\pm$0.0861 & 0.4876$\pm$0.0757 & 0.5545$\pm$0.0902 & 0.6614$\pm$0.0283 \\
\midrule

\multirow{4}{*}{CONCH~\cite{lu2024visual}}
&Adapter~\cite{gao2024clip} & 0.0114$\pm$0.0000 & \textbf{0.8240}$\pm$0.0147 & 0.7493$\pm$0.0601 & 0.6818$\pm$0.0787 & \textbf{0.7192}$\pm$0.0404 & \textbf{0.7987}$\pm$0.0322 & \textbf{0.8731}$\pm$0.0145 \\
&Classifier~\cite{ioffe2015batch} & 0.0114$\pm$0.0000 & 0.0191$\pm$0.0208 & 0.0191$\pm$0.0164 & 0.0681$\pm$0.0640 & 0.0503$\pm$0.1048 & 0.1979$\pm$0.2749 & 0.5042$\pm$0.2995 \\
&LoRA~\cite{hu2022lora} & 0.0114$\pm$0.0000 & 0.8208$\pm$0.0018 & \textbf{0.8201}$\pm$0.0036 & \textbf{0.7858}$\pm$0.0168 & 0.6286$\pm$0.0259 & 0.6797$\pm$0.0277 & 0.7864$\pm$0.0144 \\
&Vanilla~\cite{srinivasan2024comparative} & 0.0114$\pm$0.0000 & 0.5468$\pm$0.0561 & 0.5566$\pm$0.0605 & 0.6510$\pm$0.0941 & 0.6797$\pm$0.0552 & 0.7877$\pm$0.0592 & 0.8298$\pm$0.0438 \\
\midrule

\multirow{4}{*}{PLIP~\cite{huang2023visual}}
&Adapter~\cite{gao2024clip} & 0.0016$\pm$0.0000 & 0.1646$\pm$0.0216 & 0.1466$\pm$0.0315 & 0.1418$\pm$0.0268 & 0.3196$\pm$0.0546 & 0.4138$\pm$0.0739 & 0.4688$\pm$0.0348 \\
&Classifier~\cite{ioffe2015batch} & 0.0016$\pm$0.0000 & 0.1031$\pm$0.0937 & 0.1439$\pm$0.1589 & 0.2292$\pm$0.2142 & 0.3272$\pm$0.1777 & 0.4813$\pm$0.1514 & 0.4694$\pm$0.1190 \\
&LoRA~\cite{hu2022lora} & 0.0016$\pm$0.0000 & 0.2311$\pm$0.0583 & 0.1814$\pm$0.0509 & 0.1813$\pm$0.0398 & 0.3407$\pm$0.0591 & 0.4850$\pm$0.0384 & 0.5315$\pm$0.0222 \\
&Vanilla~\cite{srinivasan2024comparative} & 0.0016$\pm$0.0000 & 0.3785$\pm$0.1417 & 0.3038$\pm$0.0939 & 0.3535$\pm$0.0609 & 0.4180$\pm$0.0727 & 0.4564$\pm$0.0483 & 0.5517$\pm$0.0525 \\
\midrule

\multirow{1}{*}{ResNet-18~\cite{he2016deep}}
&Supervised Learning & \textbf{0.0914}$\pm$0.0000 & 0.0885$\pm$0.0004 & 0.1355$\pm$0.0016 & 0.1355$\pm$0.0350 & 0.0733$\pm$0.0142 & 0.1057$\pm$0.0397 & 0.0521$\pm$0.0057 \\
\midrule

\multirow{1}{*}{SSL~\cite{ciga2022self}}
&Supervised Learning & 0.0481$\pm$0.0000 & 0.0475$\pm$0.0002 & 0.0896$\pm$0.0341 & 0.0266$\pm$0.0056 & 0.0213$\pm$0.0047 & 0.0358$\pm$0.0165 & 0.0244$\pm$0.0070 \\

\bottomrule
\end{tabular}
\end{adjustbox}

\caption{\textbf{ECE and boundary margin on Cornell.} ECE (Top-1) and boundary margin mean scores (Mean$\pm$SD) across shots $\{0,1,2,4,8,16,32\}$. Column-wise best values are bolded (lower is better for ECE; higher is better for boundary margin).}
\label{tab:ece_margin_cornell}
\end{table*}

\section{Results}
\subsection{Model Validation}
\subsubsection{Classification Performance}
Tables~\ref{tab:table_1} and~\ref{tab:table_2} show a shared supervision trend across Cornell and Vanderbilt datasets: performance improves as shots increase, and AUC changes are the most consistent among the three metrics. In both ROC-AUC plots from Figs.~\ref{fig:ROC_AUC} and~\ref{fig:vandy_roc_auc}, curves progressively shift toward the upper-left region. At higher shots, the strongest configurations are dominated by CONCH-based models, with CONCH with Vanilla leading most high-shot endpoints and CONCH with Classifier giving the top AUC on Cornell.

Across backbones, CONCH shows the strongest medium-to-high-shot behavior on both datasets. Cornell presents a low-shot regime where PLIP is competitive, but CONCH becomes dominant from moderate supervision onward. On the Vanderbilt dataset, CONCH remains the most competitive backbone through most shot settings, while CLIP and PLIP improve more gradually and stay behind at high shots.
Across adaptation strategies, Vanilla exhibits the most persistent cumulative gains and most frequently attains the best medium/high-shot results, especially for Accuracy and F1. Adapter and LoRA are generally intermediate, with LoRA often narrowing the gap at higher shots. Classifier tuning is the most dataset-dependent strategy: it is strong for CONCH-AUC on Cornell, but it is consistently weak on the Vanderbilt dataset across backbones.
The clearest cross-dataset split is therefore observed in Classifier tuning, which performs competitively on Cornell but degrades on the Vanderbilt dataset. A second split is in improvement breadth: Cornell shows broader gains across multiple backbone-strategy combinations, whereas the Vanderbilt dataset concentrates its strongest gains in CONCH with Vanilla adaptation.
Qualitatively, the Cornell and Vanderbilt boxplots show a shared supervision trend: as shots increase, AUC medians generally move upward and the distributions become more concentrated, with Vanilla frequently yielding the strongest high-shot profiles (Figs.~\ref{fig:boxplot} and~\ref{fig:vandy_boxplot}). In Cornell, method trajectories are comparatively aligned, and backbone-level differences are present but moderate. By class, GSG and Viable tend to show higher and tighter distributions, whereas Ischemic and SSG remain broader with longer whiskers, indicating greater classification difficulty.

In the Vanderbilt dataset, the most salient pattern is stronger separation across adaptation strategies. Vanilla, and in several settings LoRA, show clearer upward median shifts and tighter high-shot boxes, while Classifier tuning remains consistently lower and more dispersed across backbones and classes. Relative to Cornell, this method-level divergence is substantially larger in the Vanderbilt dataset and becomes the dominant qualitative feature of the boxplot results.

\subsubsection{Cross-Multimodalities Comparison}
As shown in Figs.~\ref{fig:align_sim_result} and~\ref{fig:vandy_alignment_sim_gap}, Cornell and Vanderbilt datasets exhibit the same directional pattern in cross-modal metrics. With increasing supervision, Alignment generally rises, and Similarity Gap shifts from near-zero to clearly positive values. Across backbones, Adapter most often gives the highest Alignment, whereas Vanilla most often gives the largest Similarity Gap, with LoRA typically in between.

The main cross-dataset difference is therefore the strength of these trends. Cornell shows steeper increases and clearer method separation, especially at medium-to-high shots and most visibly on CONCH. Vanderbilt follows the same ordering but with flatter trajectories and more overlap across methods in several settings. In short, the two datasets share a consistent trend structure, while differing in effect magnitude.

Consistent with the alignment/similarity trends, the UMAP projections (Figs.~\ref{fig:umap} and~\ref{fig:vandy_kde}) show progressively improved structure with increasing shots: image clusters become more compact, class-wise regions become more separable, and text anchors move closer to their corresponding image groups in the stronger settings. On Cornell, this refinement is most evident for CONCH with Vanilla, where high-shot embeddings form the clearest modality-consistent organization. On Vanderbilt, the same directional improvement is present, but Adapter and Vanilla remain closer to each other in visual quality, with less pronounced separation between the two.

The quantitative geometry metrics follow the same pattern (Tables~\ref{tab:tabel2} and~\ref{tab:distance_vandy_run9}). In Cornell, high-shot CONCH with Vanilla most clearly combines lower Intra-Class Distance with higher Silhouette, matching the cleaner UMAP layout. In Vanderbilt, both Adapter and Vanilla improve the geometry, but their gap is modest, again reflecting weaker method separation than in Cornell. Across both datasets, Classifier tends to show less coherent geometric behavior, which is also visible as less stable cluster structure in UMAP.

\subsubsection{Confidence Calibration and Decision Boundary Margin Analysis}

We additionally report ECE and boundary margin (Tables~\ref{tab:ece_margin_cornell} and~\ref{tab:ece_margin_vandy}) to characterize confidence quality and decision boundary behavior. Lower ECE indicates better calibration, while higher margin indicates clearer class separation.

On Cornell, metrics are broadly aligned: increasing supervision generally improves classification quality, calibration, and margin in stronger settings, particularly for high-shot CONCH and Vanilla adaptation, whereas the extreme low-shot regime shows the main discrepancy, with some settings remaining low-performing despite producing high-confidence and poorly calibrated predictions, as indicated by ECE.

On Vanderbilt, the trend is more heterogeneous across methods. Although classification improves at higher shots for the better adaptation strategies, ECE often remains comparatively high, and method-level differences become more pronounced with supervision. The clearest mismatch is concentrated in Classifier tuning, where confidence-related signals remain strong while classification quality stays weak across multiple backbones.

Cross-dataset divergence is therefore most evident along two axes: strategy and shot regime. Strategy-wise, Classifier tuning is the largest split between datasets; shot-wise, Cornell mismatches are mainly early-shot, whereas Vanderbilt mismatches are more persistent in medium-to-high-shot settings.

\subsection{Clinical Validation}
\subsubsection{Expert Label Audit}
To assess label reliability from a clinical perspective, we conducted an expert label audit on a balanced subset of 500 Cornell patches (100 per class). The class-wise summary is reported in Table~\ref{tab:doctor_label_recall_kappa}. When agreement was computed jointly across all 500 reviewed patches and all five classes, the overall multiclass Fleiss' $\kappa$ was 0.762, indicating substantial agreement. This overall value is distinct from the table's Average one-vs-rest $\kappa$ entry (0.7615), which is a macro-average across classes rather than a single multiclass agreement estimate. Among the 500 reviewed patches, 365 (73.0\%) received identical labels from all three raters, and 489 (97.8\%) received the same label from at least two raters.

Agreement was not uniform across classes. Global Glomerulosclerosis showed near-perfect inter-rater agreement, while Ischemic Glomeruli showed the lowest agreement. Viable Glomeruli had high class-wise physician recall/sensitivity but lower one-vs-rest inter-rater $\kappa$, indicating that per-class recognition and agreement do not always move together. Overall, the expert audit supports the reliability and clinical plausibility of the benchmark labels while also showing that some classes are more challenging than others.
We focused this audit on the Cornell subset because Cornell serves as the primary fine-grained glomerular subtype benchmark in this study, and its five classes represent closely related morphologic subtypes with known boundary ambiguity, particularly across ischemic change, segmental sclerosis, and global sclerosis.
\begin{table*}[ht]
\centering

\begin{adjustbox}{max width=\textwidth}
\begin{tabular}{lcccccc}
\toprule
Metric & Atubular Glomeruli & Global Glomerulosclerosis & Ischemic Glomeruli & Segmental Glomerulosclerosis & Viable Glomeruli & Average \\
\midrule

\textbf{Doctor 1} & \textbf{0.6900} & \textbf{1.0000} & \textbf{0.9700} & \textbf{0.8000} & \textbf{0.9500} & \textbf{0.8820} \\
\textbf{Doctor 2} & \textbf{0.6900} & \textbf{1.0000} & \textbf{0.4100} & \textbf{0.7600} & \textbf{0.9500} & \textbf{0.7620} \\
\textbf{Doctor 3} & \textbf{0.7300} & \textbf{1.0000} & \textbf{0.7100} & \textbf{0.9100} & \textbf{0.9000} & \textbf{0.8500} \\
\midrule

\textbf{One-vs-rest Fleiss' $\kappa$} & \textbf{0.8419} & \textbf{0.9835} & \textbf{0.5497} & \textbf{0.7422} & \textbf{0.6900} & \textbf{0.7615} \\
\bottomrule
\end{tabular}
\end{adjustbox}
\caption{\textbf{Human pathologist performance baseline on a blinded glomeruli subset.} Class-wise pathologist recall and inter-pathologist agreement are reported to contextualize model performance. Recall is computed within each ground-truth class. Fleiss' $\kappa$ is computed per class using a one-vs-rest formulation; the Average entry for this row is the macro-average of the five one-vs-rest Fleiss' $\kappa$ values.
}
\label{tab:doctor_label_recall_kappa}
\end{table*}

\subsection{Discussion}
Across both cohorts, increasing supervision consistently improves downstream discrimination. At the same time, the gain profile differs by dataset: Cornell shows steeper and more stable improvements, whereas Vanderbilt follows the same direction with smaller effect size and stronger method sensitivity. This overall pattern indicates that few-shot scaling is reproducible across centers, but its practical benefit still depends on backbone and adaptation choice under dataset-specific distributions.

From the backbone perspective, pathology-specialized models outperform general CLIP, with CONCH showing the strongest high-shot ceiling. The advantage of CONCH/PLIP over CLIP is consistent with domain-informed pretraining: pathology-native representations start from a more relevant semantic prior and require fewer labels to reach useful discrimination. The further separation between CONCH and PLIP at higher shots likely reflects architectural and objective-level differences in multimodal pretraining, which become more visible once supervision is sufficient to exploit model capacity. In practical terms, this suggests a two-stage pattern in which domain knowledge dominates early-shot utility, while representational flexibility determines high-shot headroom.

From the adaptation perspective, Vanilla is the most consistently effective at medium-to-high shots, while Classifier tuning is the least robust. Vanilla fine-tuning can jointly reshape feature geometry and decision boundaries rather than only reweighting a fixed embedding space. Adapter and LoRA provide useful but more constrained updates, and they often improve progressively with additional labels. By contrast, Classifier tuning is highly conditional: it can appear competitive when frozen features are already well aligned, but it degrades rapidly when representation shift is non-trivial, producing unstable behavior across datasets and shot regimes.

The cross-modal results add an important complementary view. Adapter tends to maximize alignment, whereas Vanilla more often enlarges the similarity gap and improves geometric separability. This difference is expected because alignment and similarity gap are complementary rather than interchangeable: higher alignment mainly reflects stronger positive-pair attraction, while a larger similarity gap requires simultaneous separation from mismatched classes. Accordingly, Adapter can lead alignment without always yielding the best classification, whereas Vanilla is more consistently associated with cleaner UMAP geometry (lower intra-class distance and higher silhouette in the strongest settings). Again, Cornell and Vanderbilt show the same direction but different intensity, with clearer method separation on Cornell.

Calibration and boundary analyses further refine this picture. The key failure mode is confident error, which is more persistent on Vanderbilt. High margin alone is not sufficient for reliability if calibration is poor. In this setting, confidence-related signals can remain strong while correctness stays low, creating a clinically relevant mismatch. In our results, this mismatch is concentrated in low-shot extremes on Cornell but persists into medium/high-shot settings on Vanderbilt for less robust strategies, indicating greater confidence fragility under external domain shift.

When model failure occurs, it is concentrated in two recurring scenarios. First, errors are most likely for glomeruli that lie along the morphological continuum from ischemic change to segmental sclerosis and then global sclerosis, where subtle structural differences make discrimination more difficult under a single-label five-class formulation. This interpretation is consistent with the broader and less stable class-wise behavior observed for the more difficult categories, particularly Ischemic and SSG. Second, failure also arises under external domain shift, especially on Vanderbilt and for weaker adaptation strategies such as Classifier tuning, where confidence-related signals can remain strong even when classification quality is limited. In this setting, the dominant risk is not simply low confidence, but confident error, indicating that a high decision margin alone is insufficient evidence of reliable performance under cross-center variation.

These cross-center differences are consistent with, but cannot be attributed solely to, institution-specific appearance variation, including staining and scanner effects, superimposed on cohort heterogeneity. Within the two evaluated cohorts, increasing the target-cohort shot budget was associated with improved discrimination, embedding geometry, and cross-modal separation, particularly for stronger backbones and Vanilla adaptation. However, calibration fragility and method sensitivity persisted on Vanderbilt even at medium-to-high shot settings. Therefore, the present results suggest that additional in-domain supervision may help adaptation to center-specific visual variation, but they do not establish general mitigation of domain shift. Broader multi-center evaluations spanning additional scanners, staining protocols, and patient populations are needed to test this hypothesis.

The timing analysis adds a computational-efficiency perspective to the performance results in Tables~\ref{tab:timing_cornell} and~\ref{tab:timing_vandy}. Image-only baselines achieve the highest raw throughput, but their weaker discrimination shows that speed alone does not characterize overall model performance. Within VLMs, PLIP and CLIP generally run faster, whereas CONCH provides stronger high-shot discrimination at a higher inference-time cost. Adaptation effects on throughput are method- and cohort-dependent: Classifier tuning is often the fastest VLM adaptation strategy, but its weaker robustness offsets this computational advantage; Adapter tuning offers a potentially useful trade-off in some settings, although its throughput advantage over Vanilla is not consistent across cohorts. For CONCH, Adapter improves throughput relative to Vanilla on Vanderbilt but not on Cornell. For future translational evaluation, model comparison should jointly consider discrimination, calibration, and inference throughput rather than any single metric; these benchmark-level comparisons do not by themselves establish clinical utility.

Taken together, within this retrospective two-cohort benchmark, the findings support a research-oriented model-selection pattern rather than a deployment rule. Pathology-specialized backbones provided stronger low-shot initialization, CONCH with Vanilla adaptation achieved the strongest discrimination when sufficient supervision was available, and Adapter tuning may warrant consideration when throughput--performance trade-offs are important. Calibration and margin analyses should accompany discrimination metrics in subsequent validation studies, particularly because confident errors persisted under cross-center variation. These observations can inform the design of prospective evaluations but should not be interpreted as recommendations for clinical deployment or safety.

\subsection{Limitations}
This study has several limitations. First, we use a single fixed text template for class prompts across all experiments. This design improves experimental control and comparability across backbones and adaptation strategies, but it also limits the richness of textual supervision and may underestimate the contribution of prompt design and textual priors to downstream vision-language performance. Future work should evaluate richer textual inputs, including morphology-specific descriptions, clinically relevant contextual information, and more structured reasoning-oriented prompts, together with systematic prompt sensitivity analyses.

Second, our few-shot experiments use repeated random class-balanced subsets while keeping the WSI-level train, validation, and test split fixed. This design was chosen to enable controlled shot-scaling comparisons, so that differences between 1-, 2-, 4-, and higher-shot settings primarily reflect the addition of training examples rather than changes in WSI composition, staining, scanner characteristics, tissue quality, or validation-set difficulty. The fixed validation WSIs also provide a consistent basis for early stopping and hyperparameter selection across shot settings and adaptation strategies. However, this design does not fully quantify split-to-split variability, and the conclusions may still be influenced by the particular WSI partition used in this study. Repeated WSI-level resampling or cross-validation would address this complementary question more directly and should be considered in future larger-scale evaluations.

Third, we formulate glomerular morphology as a single-label five-class task. Although this setting provides a controlled benchmark, it may not fully capture the full spectrum of morphologic variation observed in practice, particularly along transitions from ischemic change to segmental sclerosis and then global sclerosis. A more clinically faithful extension would represent lesions with multi-label annotations or lesion tags, allowing overlapping findings to be modeled explicitly while improving interpretability and better reflecting pathological continuity.

Fourth, this study is retrospective and includes a limited number of cohorts and centers. Although the Vanderbilt cohort provides a cross-center evaluation under a distinct label schema and appearance distribution, the present design does not assess prospective workflow performance, human--AI interaction, or clinical safety, and it cannot establish broad generalizability across institutions, scanners, staining protocols, or patient populations. Accordingly, the throughput, calibration, and representation analyses should be interpreted as comparative research evidence rather than evidence of clinical readiness.

Finally, the expert label audit was limited to the Cornell subset. Similar annotation ambiguity may also affect Vanderbilt, especially near boundaries among disappearing, normal, obsolescent, solidified, and non-glomerular categories. However, Vanderbilt uses a different label schema from Cornell, and an equivalent blinded multi-rater re-annotation workflow was not available for the current study. Future work should extend expert-audit analysis to external cohorts using harmonized review protocols.

\section{Conclusion}
In this study, we frame fine-grained glomerular subtyping as a clinically motivated few-shot problem and use it to jointly examine supervision, adaptation strategy, and multimodal representation. We show that performance improves consistently with additional supervision, with meaningful gains already emerging from only a handful of labeled exemplars per class. Pathology-specialized VLM backbones offer a clear advantage in the extreme low-shot regime, but this gap narrows as more supervision becomes available. We also find that adaptation strategy governs not only accuracy but also how image and text features are organized. These observations indicate that multimodal alignment alone is not sufficient; how the model reshapes decision boundaries under limited labels is equally critical. Taken together, within this retrospective two-cohort setting, our results provide comparative evidence for model selection and annotation allocation under the evaluated data constraints. Alignment- and separability-based metrics may serve as candidate research indicators for future prospective multi-center validation, but they do not by themselves establish clinical readiness.


\acks{This research was supported by the WCM Radiology AIMI Fellowship, WCM CTSC 2027 Pilot Award, NIH 1R01HL174863-01A1(Sabuncu), and 1U54DK144866-01(Sabuncu).}

%
\ethics{The work follows appropriate ethical standards in conducting research and writing the manuscript, following all applicable laws and regulations regarding treatment of animals or human subjects.}

\coi{We declare we don't have conflicts of interest.}

\data{In accordance with institutional regulations and in order to safeguard patient privacy and confidentiality, the datasets analyzed during this study cannot be shared and are not publicly available.}

\appendix
\setcounter{figure}{0}
\renewcommand{\thefigure}{S\arabic{figure}}

\setcounter{table}{0}
\renewcommand{\thetable}{S\arabic{table}}

\setcounter{equation}{0}
\renewcommand{\theequation}{S\arabic{equation}}
\providecommand{\theHequation}{}
\renewcommand{\theHequation}{S\arabic{equation}}

\section{Detailed Mathematical Definitions}
\label{app:metric_definitions}
This appendix provides the detailed standard formulations underlying the compact final definitions retained in the main text.

\subsection{Training Objectives}
For a paired image $x_i$ and class prompt $y_i$, let $f_\theta$ and $g_\phi$ denote the image and text encoders, respectively, and let $\tau$ denote the temperature. The normalized embeddings and temperature-scaled similarities are
\begin{equation}
\mathbf{v}_i=\frac{f_\theta(x_i)}{\lVert f_\theta(x_i)\rVert},
\qquad
\mathbf{t}_i=\frac{g_\phi(y_i)}{\lVert g_\phi(y_i)\rVert},
\qquad
s_{ij}=\frac{\mathbf{v}_i^\top\mathbf{t}_j}{\tau}.
\label{eq:sup_itc_similarity}
\end{equation}
The two directional losses and their symmetric average are
\begin{equation}
\begin{aligned}
\mathcal{L}_{i\to t}
&=-\frac{1}{N}\sum_{i=1}^{N}
\log\frac{\exp(s_{ii})}{\sum_{j=1}^{N}\exp(s_{ij})},\\
\mathcal{L}_{t\to i}
&=-\frac{1}{N}\sum_{i=1}^{N}
\log\frac{\exp(s_{ii})}{\sum_{j=1}^{N}\exp(s_{ji})},\\
\mathcal{L}_{\mathrm{ITC}}
&=\frac{1}{2}\left(\mathcal{L}_{i\to t}+\mathcal{L}_{t\to i}\right).
\end{aligned}
\label{eq:sup_itc_directional}
\end{equation}
For Classifier tuning, $z_{i,c}$ denotes the logit for class $c$ among $C$ classes, and the multiclass cross-entropy objective is
\begin{equation}
p_{i,c}=\frac{\exp(z_{i,c})}{\sum_{r=1}^{C}\exp(z_{i,r})},
\qquad
\mathcal{L}_{\mathrm{cls}}
=-\frac{1}{N}\sum_{i=1}^{N}\log p_{i,y_i}.
\label{eq:sup_classifier_loss}
\end{equation}

\subsection{Standard Evaluation Metrics}
For the Silhouette score computed in the original embedding space with cosine distance, let $n_c$ denote the number of samples in class $c$. The mean within-class distance $a_i$ and the minimum mean distance to another class $b_i$ are
\begin{equation}
\begin{aligned}
a_i&=\frac{1}{n_{y_i}-1}
\sum_{\substack{j\ne i\\y_j=y_i}}
d_{\cos}(\mathbf{x}_i,\mathbf{x}_j),\\
b_i&=\min_{c\ne y_i}\frac{1}{n_c}
\sum_{y_j=c}d_{\cos}(\mathbf{x}_i,\mathbf{x}_j).
\end{aligned}
\label{eq:sup_silhouette_components}
\end{equation}
The individual coefficient is $s_i=(b_i-a_i)/\max\{a_i,b_i\}$, and the dataset-level score reported in the main text is the mean of $s_i$ over all samples.

For top-1 ECE, let $\hat{y}_i=\arg\max_c p_{i,c}$ and $q_i=\max_c p_{i,c}$ denote the predicted class and confidence. For each of the 15 equal-width bins $B_m$ over $[0,1]$, bin accuracy and confidence are
\begin{equation}
\begin{aligned}
\mathrm{acc}(B_m)
&= \frac{1}{|B_m|}
\sum_{i\in B_m}
\mathbf{1}\{\hat{y}_i = y_i\}, \\
\mathrm{conf}(B_m)
&= \frac{1}{|B_m|}
\sum_{i\in B_m} q_i.
\end{aligned}
\label{eq:sup_ece_components}
\end{equation}
Empty bins contribute zero to the weighted ECE sum.

For Fleiss' $\kappa$, consider $N$ samples rated by $n$ raters into $k$ classes, and let $n_{ij}$ be the number of raters who assign sample $i$ to class $j$. The per-sample and mean observed agreement are
\begin{equation}
P_i=\frac{1}{n(n-1)}\sum_{j=1}^{k}n_{ij}(n_{ij}-1),
\qquad
\bar{P}=\frac{1}{N}\sum_{i=1}^{N}P_i.
\label{eq:sup_fleiss_observed}
\end{equation}
The marginal class proportion and chance-expected agreement are
\begin{equation}
p_j=\frac{1}{Nn}\sum_{i=1}^{N}n_{ij},
\qquad
\bar{P}_e=\sum_{j=1}^{k}p_j^2.
\label{eq:sup_fleiss_expected}
\end{equation}
Substitution of $\bar{P}$ and $\bar{P}_e$ into Eq.~\ref{eq:fleiss_kappa} yields the reported agreement statistic.

\section{Additional Results on Vanderbilt Dataset}
This appendix provides supplementary results on the Vanderbilt dataset to support the main findings and improve reproducibility.

\begin{figure*}[!t]
\centering
\includegraphics[width=\linewidth]{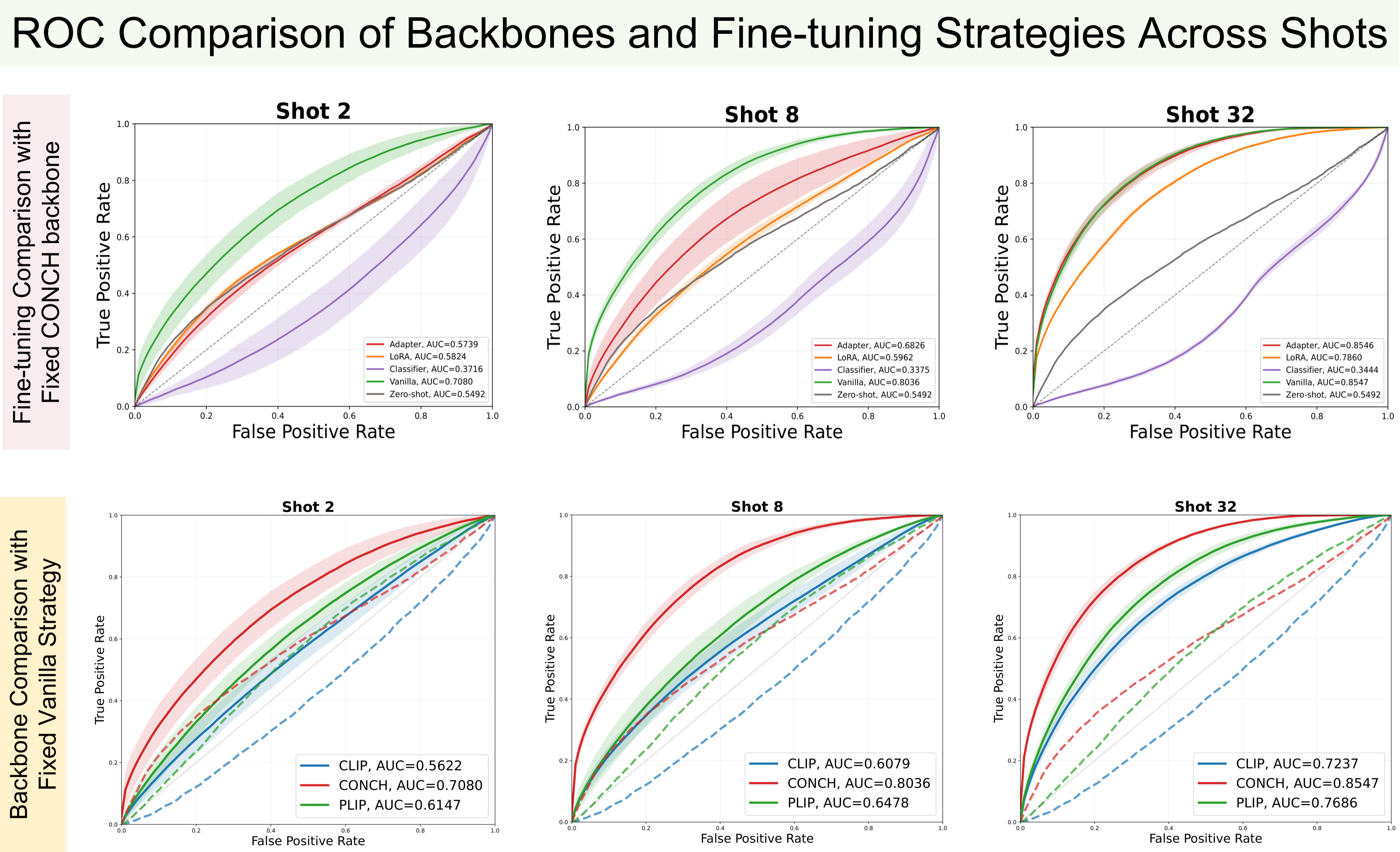}
\caption{\textbf{ROC curves across models and fine-tuning methods on Vanderbilt dataset.} We compare backbones under fixed strategies and strategies under fixed backbones by plotting TPR versus FPR across increasing shot counts.}
\label{fig:vandy_roc_auc}
\end{figure*}

\begin{figure*}[!t]
\centering
\includegraphics[width=\linewidth]{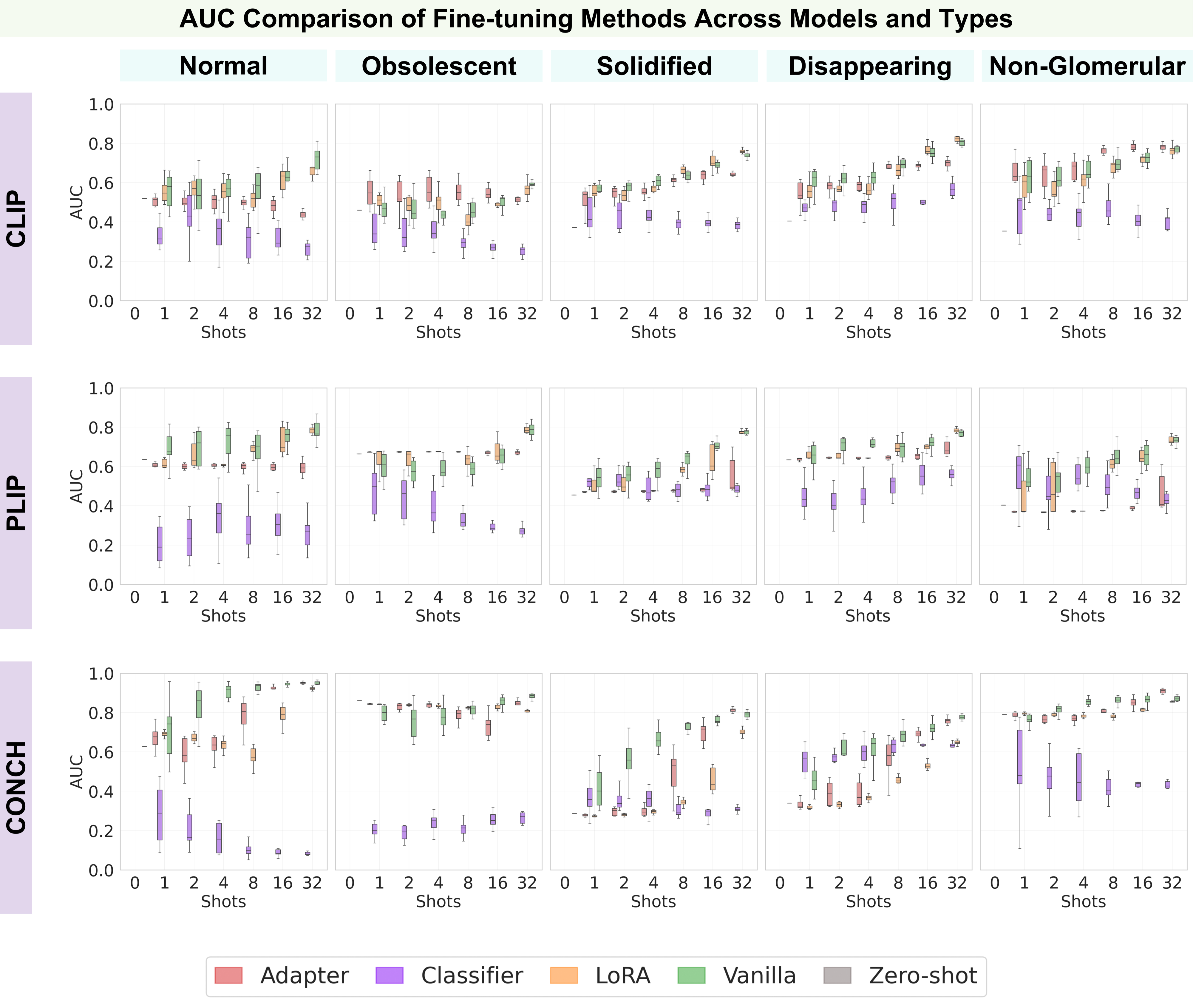}
\caption{\textbf{Discrimination across subtypes on Vanderbilt.} This figure summarizes per-class discrimination performance across 10 runs. For each glomerular subtype, we report the distribution of AUC across VLM backbones, fine-tuning strategies, and shot settings.}
\label{fig:vandy_boxplot}
\end{figure*}

\begin{figure*}[!t]
\centering
\includegraphics[width=\linewidth]{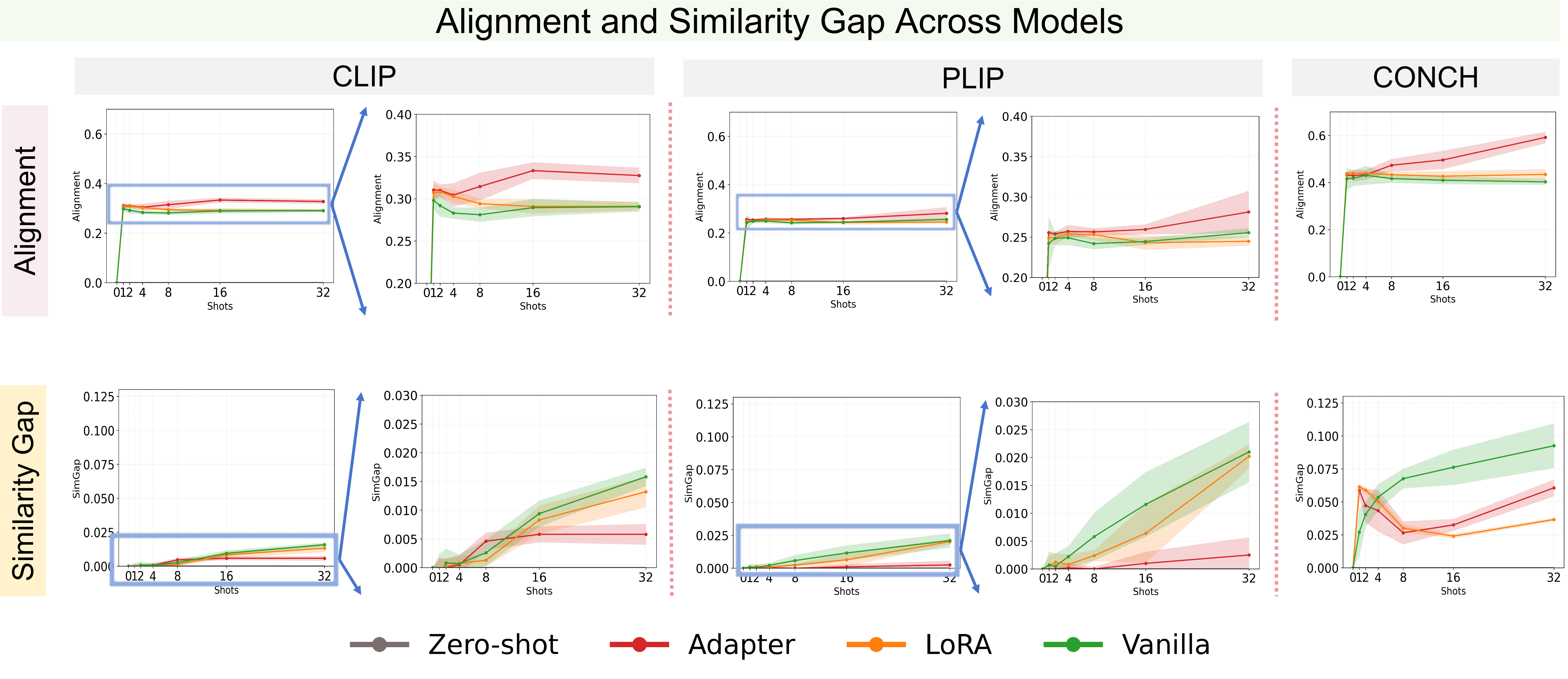}
\caption{\textbf{Alignment and similarity gap across models on Vanderbilt.} This figure shows shot-wise trajectories of Alignment and Similarity Gap for CLIP, PLIP, and CONCH under different fine-tuning strategies. Similarity Gap is defined as mean positive class-prompt similarity minus mean negative class-prompt similarity. CLIP and PLIP are shown in both full-scale and zoomed-in panels for readability.}
\label{fig:vandy_alignment_sim_gap}
\end{figure*}

\begin{figure*}[!t]
\centering
\includegraphics[width=\linewidth]{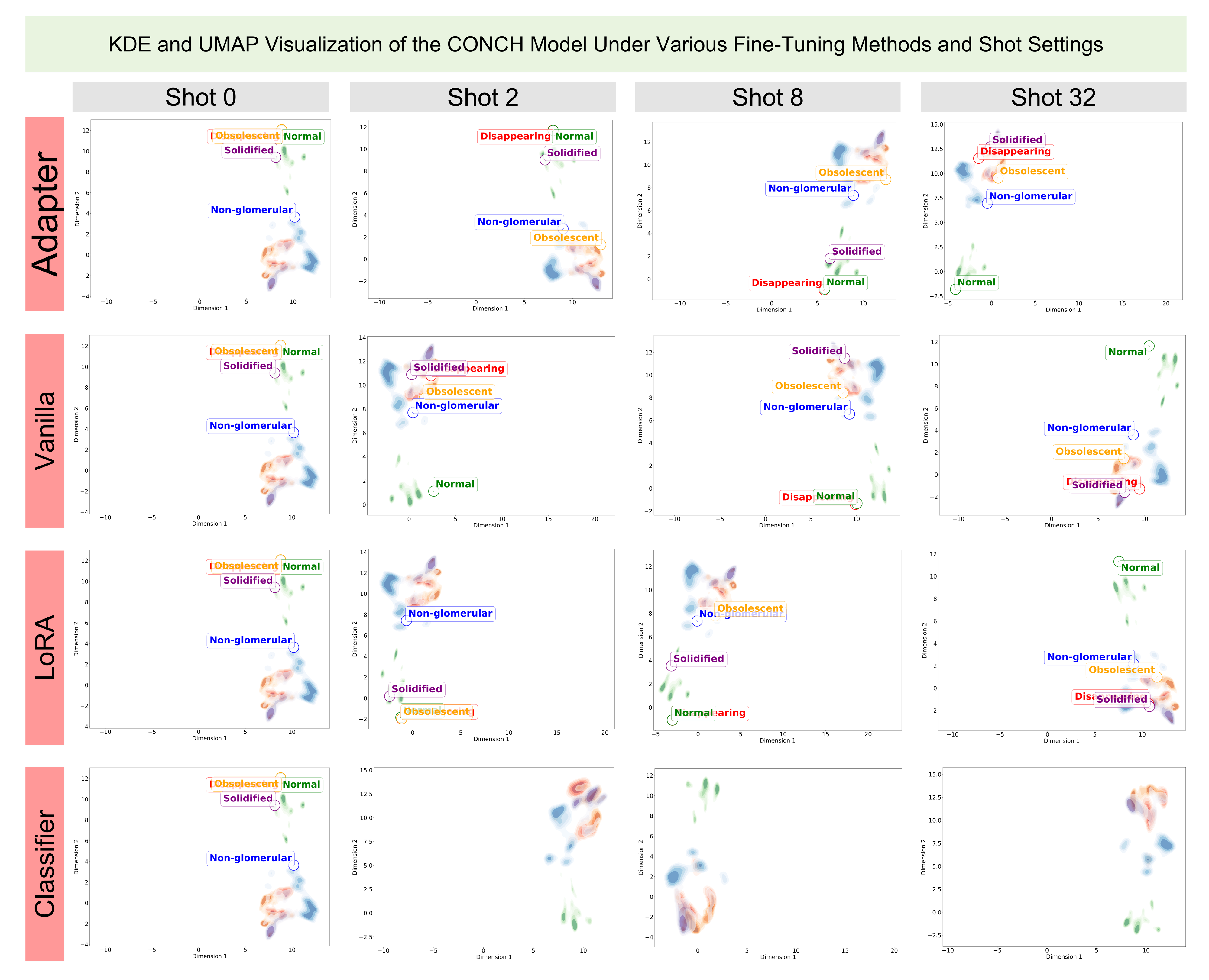}
\caption{\textbf{UMAP/KDE visualization of embeddings across shots on Vanderbilt.}
This figure shows text and image embeddings from the best CONCH run across shot settings. For Classifier tuning, text is not updated and only image-side embeddings are shown. Specifically, for Classifier tuning we visualize the penultimate hidden representation of the tuned MLP head (fed by fixed 512-D CONCH features), which changes as the head is updated. The companion Table~\ref{tab:distance_vandy_run9} reports Intra-Class Distance and Silhouette Score.}
\label{fig:vandy_kde}
\end{figure*}

\begin{table*}[!t]
\centering

\begin{adjustbox}{max width=\textwidth}
\begin{tabular}{llccccccc}
\toprule
\multirow{2}{0.4in}{Model} & Method & \multicolumn{7}{c}{Accuracy} \\[0.3em]
\cmidrule{3-9}
& Shots & 0 & 1 & 2 & 4 & 8 & 16 & 32 \\
\midrule

\multirow{4}{*}{CLIP~\cite{radford2021learning}}
&Adapter~\cite{gao2024clip} & 0.2873$\pm$0.0000 & 0.1898$\pm$0.0688 & 0.2118$\pm$0.0615 & 0.2364$\pm$0.0630 & 0.3426$\pm$0.0316 & 0.3575$\pm$0.0182 & 0.3514$\pm$0.0174 \\
&Classifier~\cite{ioffe2015batch} & 0.2873$\pm$0.0000 & 0.1456$\pm$0.0739 & 0.1091$\pm$0.0450 & 0.1079$\pm$0.0201 & 0.0935$\pm$0.0275 & 0.0796$\pm$0.0176 & 0.0691$\pm$0.0070 \\
&LoRA~\cite{hu2022lora} & 0.2873$\pm$0.0000 & 0.2531$\pm$0.0667 & 0.2683$\pm$0.0577 & 0.2820$\pm$0.0703 & 0.2667$\pm$0.0400 & 0.3407$\pm$0.0364 & 0.4025$\pm$0.0283 \\
&Vanilla~\cite{srinivasan2024comparative} & 0.2873$\pm$0.0000 & 0.2505$\pm$0.0851 & 0.2924$\pm$0.0668 & 0.2699$\pm$0.0658 & 0.3245$\pm$0.0460 & 0.3266$\pm$0.0404 & 0.4023$\pm$0.0267 \\
\midrule

\multirow{4}{*}{CONCH~\cite{lu2024visual}}
&Adapter~\cite{gao2024clip} & \textbf{0.3281}$\pm$0.0000 & 0.3576$\pm$0.0386 & 0.2977$\pm$0.0677 & 0.3168$\pm$0.0518 & 0.3639$\pm$0.0520 & 0.4173$\pm$0.0291 & 0.5053$\pm$0.0321 \\
&Classifier~\cite{ioffe2015batch} & \textbf{0.3281}$\pm$0.0000 & 0.1283$\pm$0.0608 & 0.0812$\pm$0.0380 & 0.1461$\pm$0.1012 & 0.0574$\pm$0.0251 & 0.0504$\pm$0.0112 & 0.0475$\pm$0.0088 \\
&LoRA~\cite{hu2022lora} & \textbf{0.3281}$\pm$0.0000 & \textbf{0.3666}$\pm$0.0065 & 0.3657$\pm$0.0114 & 0.3421$\pm$0.0274 & 0.2944$\pm$0.0317 & 0.3515$\pm$0.0297 & 0.4421$\pm$0.0067 \\
&Vanilla~\cite{srinivasan2024comparative} & \textbf{0.3281}$\pm$0.0000 & 0.2856$\pm$0.0722 & \textbf{0.3867}$\pm$0.0712 & \textbf{0.4416}$\pm$0.0542 & \textbf{0.4689}$\pm$0.0267 & \textbf{0.4623}$\pm$0.0301 & \textbf{0.5111}$\pm$0.0441 \\
\midrule

\multirow{4}{*}{PLIP~\cite{huang2023visual}}
&Adapter~\cite{gao2024clip} & 0.1773$\pm$0.0000 & 0.1691$\pm$0.0139 & 0.1681$\pm$0.0278 & 0.1776$\pm$0.0352 & 0.1661$\pm$0.0127 & 0.1944$\pm$0.0455 & 0.2124$\pm$0.0331 \\
&Classifier~\cite{ioffe2015batch} & 0.1773$\pm$0.0000 & 0.1414$\pm$0.0560 & 0.1326$\pm$0.0497 & 0.1324$\pm$0.0317 & 0.1063$\pm$0.0221 & 0.1067$\pm$0.0313 & 0.0989$\pm$0.0230 \\
&LoRA~\cite{hu2022lora} & 0.1773$\pm$0.0000 & 0.1981$\pm$0.0475 & 0.2380$\pm$0.0816 & 0.2080$\pm$0.0949 & 0.3262$\pm$0.0367 & 0.3227$\pm$0.0457 & 0.3992$\pm$0.0280 \\
&Vanilla~\cite{srinivasan2024comparative} & 0.1773$\pm$0.0000 & 0.2314$\pm$0.0574 & 0.2403$\pm$0.0700 & 0.3096$\pm$0.0399 & 0.3301$\pm$0.0667 & 0.3289$\pm$0.0478 & 0.3852$\pm$0.0243 \\
\midrule

\multirow{1}{*}{ResNet-18~\cite{he2016deep}}
&Supervised Learning & 0.1560$\pm$0.0000 & 0.3486$\pm$0.0002 & 0.2481$\pm$0.0006 & 0.2196$\pm$0.0772 & 0.1355$\pm$0.0815 & 0.2261$\pm$0.0824 & 0.2192$\pm$0.0641 \\
\midrule

\multirow{1}{*}{SSL~\cite{ciga2022self}}
&Supervised Learning & 0.2504$\pm$0.0000 & 0.2529$\pm$0.0014 & 0.0650$\pm$0.0195 & 0.2145$\pm$0.0749 & 0.2430$\pm$0.0880 & 0.2072$\pm$0.0960 & 0.1568$\pm$0.0900 \\

\bottomrule
\end{tabular}
\end{adjustbox}

\vspace{1em}

\begin{adjustbox}{max width=\textwidth}
\begin{tabular}{llccccccc}
\toprule
\multirow{2}{0.4in}{Model} & Method & \multicolumn{7}{c}{AUC} \\[0.3em]
\cmidrule{3-9}
& Shots & 0 & 1 & 2 & 4 & 8 & 16 & 32 \\
\midrule

\multirow{4}{*}{CLIP~\cite{radford2021learning}}
&Adapter~\cite{gao2024clip} & 0.4096$\pm$0.0000 & 0.5454$\pm$0.0452 & 0.5519$\pm$0.0496 & 0.5658$\pm$0.0458 & 0.6215$\pm$0.0215 & 0.6243$\pm$0.0154 & 0.6149$\pm$0.0069 \\
&Classifier~\cite{ioffe2015batch} & 0.4096$\pm$0.0000 & 0.4145$\pm$0.0594 & 0.4289$\pm$0.0407 & 0.4094$\pm$0.0407 & 0.3914$\pm$0.0275 & 0.3796$\pm$0.0335 & 0.3759$\pm$0.0161 \\
&LoRA~\cite{hu2022lora} & 0.4096$\pm$0.0000 & 0.5526$\pm$0.0285 & 0.5374$\pm$0.0340 & 0.5602$\pm$0.0291 & 0.5855$\pm$0.0258 & 0.6607$\pm$0.0283 & 0.7168$\pm$0.0189 \\
&Vanilla~\cite{srinivasan2024comparative} & 0.4096$\pm$0.0000 & 0.5630$\pm$0.0349 & 0.5622$\pm$0.0335 & 0.5769$\pm$0.0278 & 0.6079$\pm$0.0280 & 0.6581$\pm$0.0280 & 0.7237$\pm$0.0139 \\
\midrule

\multirow{4}{*}{CONCH~\cite{lu2024visual}}
&Adapter~\cite{gao2024clip} & 0.5492$\pm$0.0000 & 0.5815$\pm$0.0070 & 0.5739$\pm$0.0081 & 0.5875$\pm$0.0225 & 0.6826$\pm$0.0596 & 0.7797$\pm$0.0186 & 0.8546$\pm$0.0108 \\
&Classifier~\cite{ioffe2015batch} & 0.5492$\pm$0.0000 & 0.3852$\pm$0.0722 & 0.3716$\pm$0.0573 & 0.3818$\pm$0.0639 & 0.3375$\pm$0.0316 & 0.3393$\pm$0.0179 & 0.3444$\pm$0.0072 \\
&LoRA~\cite{hu2022lora} & 0.5492$\pm$0.0000 & 0.5840$\pm$0.0016 & 0.5824$\pm$0.0028 & 0.5806$\pm$0.0063 & 0.5962$\pm$0.0106 & 0.6828$\pm$0.0146 & 0.7860$\pm$0.0039 \\
&Vanilla~\cite{srinivasan2024comparative} & 0.5492$\pm$0.0000 & \textbf{0.6252}$\pm$0.0460 & \textbf{0.7080}$\pm$0.0441 & \textbf{0.7560}$\pm$0.0294 & \textbf{0.8036}$\pm$0.0148 & \textbf{0.8289}$\pm$0.0175 & \textbf{0.8547}$\pm$0.0085 \\
\midrule

\multirow{4}{*}{PLIP~\cite{huang2023visual}}
&Adapter~\cite{gao2024clip} & 0.5502$\pm$0.0000 & 0.5542$\pm$0.0101 & 0.5577$\pm$0.0157 & 0.5626$\pm$0.0220 & 0.5579$\pm$0.0104 & 0.5674$\pm$0.0193 & 0.5928$\pm$0.0476 \\
&Classifier~\cite{ioffe2015batch} & 0.5502$\pm$0.0000 & 0.4485$\pm$0.0649 & 0.4156$\pm$0.0367 & 0.4374$\pm$0.0462 & 0.4199$\pm$0.0337 & 0.4201$\pm$0.0269 & 0.4043$\pm$0.0215 \\
&LoRA~\cite{hu2022lora} & 0.5502$\pm$0.0000 & 0.5749$\pm$0.0330 & 0.5905$\pm$0.0389 & 0.5781$\pm$0.0474 & 0.6412$\pm$0.0177 & 0.6766$\pm$0.0464 & 0.7719$\pm$0.0095 \\
&Vanilla~\cite{srinivasan2024comparative} & 0.5502$\pm$0.0000 & 0.6028$\pm$0.0264 & 0.6147$\pm$0.0287 & 0.6370$\pm$0.0269 & 0.6478$\pm$0.0345 & 0.6962$\pm$0.0233 & 0.7686$\pm$0.0146 \\
\midrule

\multirow{1}{*}{ResNet-18~\cite{he2016deep}}
&Supervised Learning & 0.5065$\pm$0.0000 & 0.5681$\pm$0.0002 & 0.4338$\pm$0.0007 & 0.5167$\pm$0.0544 & 0.4761$\pm$0.0364 & 0.5245$\pm$0.0506 & 0.5262$\pm$0.0605 \\
\midrule

\multirow{1}{*}{SSL~\cite{ciga2022self}}
&Supervised Learning & \textbf{0.5588}$\pm$0.0000 & 0.5566$\pm$0.0005 & 0.4917$\pm$0.0209 & 0.4872$\pm$0.0385 & 0.4971$\pm$0.0677 & 0.5346$\pm$0.0645 & 0.4524$\pm$0.0588 \\

\bottomrule
\end{tabular}
\end{adjustbox}

\vspace{1em}

\begin{adjustbox}{max width=\textwidth}
\begin{tabular}{llccccccc}
\toprule
\multirow{2}{0.4in}{Model} & Method & \multicolumn{7}{c}{F1} \\[0.3em]
\cmidrule{3-9}
& Shots & 0 & 1 & 2 & 4 & 8 & 16 & 32 \\
\midrule

\multirow{4}{*}{CLIP~\cite{radford2021learning}}
&Adapter~\cite{gao2024clip} & 0.1049$\pm$0.0000 & 0.1362$\pm$0.0438 & 0.1519$\pm$0.0314 & 0.1666$\pm$0.0353 & 0.2392$\pm$0.0159 & 0.2468$\pm$0.0101 & 0.2392$\pm$0.0099 \\
&Classifier~\cite{ioffe2015batch} & 0.1049$\pm$0.0000 & 0.0987$\pm$0.0420 & 0.0774$\pm$0.0318 & 0.0894$\pm$0.0171 & 0.0801$\pm$0.0179 & 0.0711$\pm$0.0149 & 0.0620$\pm$0.0052 \\
&LoRA~\cite{hu2022lora} & 0.1049$\pm$0.0000 & 0.1796$\pm$0.0426 & 0.1879$\pm$0.0265 & 0.2026$\pm$0.0322 & 0.2182$\pm$0.0271 & 0.2676$\pm$0.0238 & 0.3131$\pm$0.0212 \\
&Vanilla~\cite{srinivasan2024comparative} & 0.1049$\pm$0.0000 & 0.1706$\pm$0.0462 & 0.1878$\pm$0.0372 & 0.2132$\pm$0.0405 & 0.2528$\pm$0.0270 & 0.2608$\pm$0.0363 & 0.3096$\pm$0.0223 \\
\midrule

\multirow{4}{*}{CONCH~\cite{lu2024visual}}
&Adapter~\cite{gao2024clip} & \textbf{0.1894}$\pm$0.0000 & \textbf{0.2331}$\pm$0.0202 & 0.2006$\pm$0.0396 & 0.2253$\pm$0.0272 & 0.2762$\pm$0.0394 & 0.3351$\pm$0.0224 & 0.4111$\pm$0.0240 \\
&Classifier~\cite{ioffe2015batch} & \textbf{0.1894}$\pm$0.0000 & 0.0833$\pm$0.0332 & 0.0602$\pm$0.0270 & 0.0852$\pm$0.0357 & 0.0547$\pm$0.0161 & 0.0495$\pm$0.0117 & 0.0492$\pm$0.0068 \\
&LoRA~\cite{hu2022lora} & \textbf{0.1894}$\pm$0.0000 & 0.2298$\pm$0.0079 & 0.2417$\pm$0.0103 & 0.2416$\pm$0.0121 & 0.2197$\pm$0.0184 & 0.2783$\pm$0.0187 & 0.3463$\pm$0.0067 \\
&Vanilla~\cite{srinivasan2024comparative} & \textbf{0.1894}$\pm$0.0000 & 0.2242$\pm$0.0450 & \textbf{0.3017}$\pm$0.0556 & \textbf{0.3552}$\pm$0.0469 & \textbf{0.3795}$\pm$0.0223 & \textbf{0.3750}$\pm$0.0180 & \textbf{0.4118}$\pm$0.0315 \\
\midrule

\multirow{4}{*}{PLIP~\cite{huang2023visual}}
&Adapter~\cite{gao2024clip} & 0.1361$\pm$0.0000 & 0.1369$\pm$0.0091 & 0.1377$\pm$0.0216 & 0.1453$\pm$0.0239 & 0.1391$\pm$0.0113 & 0.1548$\pm$0.0258 & 0.1749$\pm$0.0222 \\
&Classifier~\cite{ioffe2015batch} & 0.1361$\pm$0.0000 & 0.1136$\pm$0.0376 & 0.0988$\pm$0.0207 & 0.1080$\pm$0.0182 & 0.0918$\pm$0.0150 & 0.0928$\pm$0.0226 & 0.0815$\pm$0.0099 \\
&LoRA~\cite{hu2022lora} & 0.1361$\pm$0.0000 & 0.1592$\pm$0.0326 & 0.1899$\pm$0.0638 & 0.1678$\pm$0.0726 & 0.2628$\pm$0.0249 & 0.2698$\pm$0.0325 & 0.3179$\pm$0.0224 \\
&Vanilla~\cite{srinivasan2024comparative} & 0.1361$\pm$0.0000 & 0.1792$\pm$0.0403 & 0.1973$\pm$0.0472 & 0.2305$\pm$0.0291 & 0.2469$\pm$0.0508 & 0.2702$\pm$0.0313 & 0.3013$\pm$0.0195 \\
\midrule

\multirow{1}{*}{ResNet-18~\cite{he2016deep}}
&Supervised Learning & 0.0792$\pm$0.0000 & 0.1686$\pm$0.0001 & 0.0850$\pm$0.0001 & 0.1129$\pm$0.0310 & 0.0924$\pm$0.0433 & 0.1335$\pm$0.0413 & 0.1696$\pm$0.0428 \\
\midrule

\multirow{1}{*}{SSL~\cite{ciga2022self}}
&Supervised Learning & 0.1205$\pm$0.0000 & 0.1234$\pm$0.0014 & 0.0386$\pm$0.0133 & 0.1262$\pm$0.0230 & 0.1528$\pm$0.0415 & 0.1206$\pm$0.0417 & 0.1025$\pm$0.0440 \\

\bottomrule
\end{tabular}
\end{adjustbox}

\caption{\textbf{Few-shot classification performance on Vanderbilt.}
Mean$\pm$SD Accuracy, AUC, and F1 across shot settings $\{0,1,2,4,8,16,32\}$ for all backbone and adaptation pairs.
Here, $k$-shot denotes $k$ labeled images per class used for fine-tuning, and 0-shot denotes the zero-shot baseline.
Bold indicates the best value within each shot column (higher is better).}
\label{tab:table_2}
\end{table*}

\begin{table*}[!t]
\centering

\begin{adjustbox}{max width=\textwidth}
\begin{tabular}{llccccccc}
\toprule
\multirow{2}{0.4in}{Model} & Method & \multicolumn{7}{c}{ECE (Top-1)} \\[0.3em]
\cmidrule{3-9}
& Shots & 0 & 1 & 2 & 4 & 8 & 16 & 32 \\
\midrule

\multirow{4}{*}{CLIP~\cite{radford2021learning}}
&Adapter~\cite{gao2024clip} & 0.0870$\pm$0.0000 & 0.1341$\pm$0.0581 & 0.1248$\pm$0.0411 & 0.1073$\pm$0.0427 & 0.0952$\pm$0.0318 & 0.1410$\pm$0.0332 & 0.1527$\pm$0.0423 \\
&Classifier~\cite{ioffe2015batch} & 0.0870$\pm$0.0000 & 0.2010$\pm$0.0774 & 0.2289$\pm$0.0972 & 0.3632$\pm$0.1739 & 0.4579$\pm$0.1404 & 0.5120$\pm$0.1267 & 0.5888$\pm$0.0369 \\
&LoRA~\cite{hu2022lora} & 0.0870$\pm$0.0000 & 0.0841$\pm$0.0443 & \textbf{0.0610}$\pm$0.0348 & \textbf{0.0577}$\pm$0.0339 & 0.1809$\pm$0.0313 & 0.2759$\pm$0.0410 & 0.2606$\pm$0.0323 \\
&Vanilla~\cite{srinivasan2024comparative} & 0.0870$\pm$0.0000 & 0.1920$\pm$0.1212 & 0.1451$\pm$0.0759 & 0.1143$\pm$0.0501 & 0.1928$\pm$0.1120 & 0.3420$\pm$0.0484 & 0.3088$\pm$0.0394 \\
\midrule

\multirow{4}{*}{CONCH~\cite{lu2024visual}}
&Adapter~\cite{gao2024clip} & 0.1447$\pm$0.0000 & 0.5159$\pm$0.0422 & 0.5653$\pm$0.0639 & 0.5235$\pm$0.0452 & 0.4120$\pm$0.0726 & 0.3881$\pm$0.0330 & 0.3368$\pm$0.0371 \\
&Classifier~\cite{ioffe2015batch} & 0.1447$\pm$0.0000 & 0.0958$\pm$0.0510 & 0.1492$\pm$0.0247 & 0.1432$\pm$0.0943 & 0.3790$\pm$0.1544 & 0.5958$\pm$0.1671 & 0.7206$\pm$0.0133 \\
&LoRA~\cite{hu2022lora} & 0.1447$\pm$0.0000 & 0.5180$\pm$0.0112 & 0.4996$\pm$0.0168 & 0.5042$\pm$0.0269 & 0.5205$\pm$0.0365 & 0.4185$\pm$0.0263 & 0.3632$\pm$0.0060 \\
&Vanilla~\cite{srinivasan2024comparative} & 0.1447$\pm$0.0000 & 0.4552$\pm$0.0770 & 0.3772$\pm$0.0707 & 0.3467$\pm$0.0590 & 0.3481$\pm$0.0320 & 0.3570$\pm$0.0354 & 0.3254$\pm$0.0413 \\
\midrule

\multirow{4}{*}{PLIP~\cite{huang2023visual}}
&Adapter~\cite{gao2024clip} & 0.0429$\pm$0.0000 & 0.2369$\pm$0.0313 & 0.2353$\pm$0.0539 & 0.2254$\pm$0.0434 & 0.2199$\pm$0.0338 & 0.1949$\pm$0.0752 & 0.2247$\pm$0.1242 \\
&Classifier~\cite{ioffe2015batch} & 0.0429$\pm$0.0000 & 0.2584$\pm$0.1010 & 0.2107$\pm$0.1731 & 0.3677$\pm$0.1989 & 0.5620$\pm$0.0383 & 0.5245$\pm$0.0923 & 0.5485$\pm$0.0280 \\
&LoRA~\cite{hu2022lora} & 0.0429$\pm$0.0000 & 0.2303$\pm$0.0547 & 0.1777$\pm$0.0853 & 0.2135$\pm$0.0813 & \textbf{0.0482}$\pm$0.0290 & 0.1580$\pm$0.1389 & 0.3000$\pm$0.0354 \\
&Vanilla~\cite{srinivasan2024comparative} & 0.0429$\pm$0.0000 & 0.2790$\pm$0.1210 & 0.1788$\pm$0.0723 & 0.1176$\pm$0.0618 & 0.1597$\pm$0.1122 & 0.3297$\pm$0.0766 & 0.3367$\pm$0.0341 \\
\midrule

\multirow{1}{*}{ResNet-18~\cite{he2016deep}}
&Supervised Learning & 0.2112$\pm$0.0000 & \textbf{0.0223}$\pm$0.0008 & 0.1735$\pm$0.0022 & 0.1328$\pm$0.0599 & 0.1827$\pm$0.0833 & 0.1069$\pm$0.0597 & \textbf{0.0818}$\pm$0.0326 \\
\midrule

\multirow{1}{*}{SSL~\cite{ciga2022self}}
&Supervised Learning & \textbf{0.0243}$\pm$0.0000 & 0.0295$\pm$0.0012 & 0.2131$\pm$0.0274 & 0.0757$\pm$0.0155 & 0.0818$\pm$0.0348 & \textbf{0.0847}$\pm$0.0549 & 0.1102$\pm$0.0581 \\

\bottomrule
\end{tabular}
\end{adjustbox}

\vspace{1em}

\begin{adjustbox}{max width=\textwidth}
\begin{tabular}{llccccccc}
\toprule
\multirow{2}{0.4in}{Model} & Method & \multicolumn{7}{c}{Boundary Margin (Mean)} \\[0.3em]
\cmidrule{3-9}
& Shots & 0 & 1 & 2 & 4 & 8 & 16 & 32 \\
\midrule

\multirow{4}{*}{CLIP~\cite{radford2021learning}}
&Adapter~\cite{gao2024clip} & 0.0013$\pm$0.0000 & 0.0807$\pm$0.0361 & 0.0901$\pm$0.0316 & 0.0915$\pm$0.0311 & 0.2123$\pm$0.0647 & 0.2749$\pm$0.0407 & 0.2781$\pm$0.0606 \\
&Classifier~\cite{ioffe2015batch} & 0.0013$\pm$0.0000 & 0.1139$\pm$0.0468 & 0.1166$\pm$0.0966 & 0.2580$\pm$0.1813 & 0.3326$\pm$0.1363 & 0.3833$\pm$0.1299 & 0.4605$\pm$0.0483 \\
&LoRA~\cite{hu2022lora} & 0.0013$\pm$0.0000 & 0.0844$\pm$0.0236 & 0.0793$\pm$0.0233 & 0.0787$\pm$0.0298 & 0.1989$\pm$0.0323 & 0.3947$\pm$0.0476 & 0.4592$\pm$0.0379 \\
&Vanilla~\cite{srinivasan2024comparative} & 0.0013$\pm$0.0000 & 0.1888$\pm$0.0642 & 0.1909$\pm$0.0614 & 0.1366$\pm$0.0300 & 0.2732$\pm$0.0842 & 0.4595$\pm$0.0205 & 0.5216$\pm$0.0277 \\
\midrule

\multirow{4}{*}{CONCH~\cite{lu2024visual}}
&Adapter~\cite{gao2024clip} & 0.0099$\pm$0.0000 & 0.7657$\pm$0.0305 & 0.7441$\pm$0.0368 & 0.7087$\pm$0.0649 & 0.6133$\pm$0.0537 & 0.6634$\pm$0.0224 & \textbf{0.7206}$\pm$0.0184 \\
&Classifier~\cite{ioffe2015batch} & 0.0099$\pm$0.0000 & 0.0128$\pm$0.0114 & 0.0231$\pm$0.0141 & 0.0426$\pm$0.0498 & 0.2173$\pm$0.1444 & 0.4645$\pm$0.1902 & 0.6049$\pm$0.0245 \\
&LoRA~\cite{hu2022lora} & 0.0099$\pm$0.0000 & \textbf{0.7878}$\pm$0.0162 & \textbf{0.7512}$\pm$0.0290 & \textbf{0.7145}$\pm$0.0204 & 0.6587$\pm$0.0122 & 0.6024$\pm$0.0170 & 0.6656$\pm$0.0057 \\
&Vanilla~\cite{srinivasan2024comparative} & 0.0099$\pm$0.0000 & 0.5539$\pm$0.0508 & 0.6001$\pm$0.0677 & 0.6389$\pm$0.0348 & \textbf{0.6846}$\pm$0.0295 & \textbf{0.6863}$\pm$0.0331 & 0.6957$\pm$0.0335 \\
\midrule

\multirow{4}{*}{PLIP~\cite{huang2023visual}}
&Adapter~\cite{gao2024clip} & 0.0012$\pm$0.0000 & 0.1698$\pm$0.0247 & 0.1668$\pm$0.0280 & 0.1673$\pm$0.0136 & 0.1491$\pm$0.0301 & 0.1526$\pm$0.0654 & 0.2038$\pm$0.1620 \\
&Classifier~\cite{ioffe2015batch} & 0.0012$\pm$0.0000 & 0.1728$\pm$0.0720 & 0.1204$\pm$0.1606 & 0.2904$\pm$0.1981 & 0.4717$\pm$0.0547 & 0.4235$\pm$0.1052 & 0.4406$\pm$0.0488 \\
&LoRA~\cite{hu2022lora} & 0.0012$\pm$0.0000 & 0.1875$\pm$0.0342 & 0.1742$\pm$0.0234 & 0.1730$\pm$0.0181 & 0.1215$\pm$0.0151 & 0.2423$\pm$0.1584 & 0.4967$\pm$0.0216 \\
&Vanilla~\cite{srinivasan2024comparative} & 0.0012$\pm$0.0000 & 0.2693$\pm$0.1311 & 0.1617$\pm$0.0329 & 0.1697$\pm$0.0732 & 0.2475$\pm$0.1235 & 0.4444$\pm$0.0975 & 0.5302$\pm$0.0482 \\
\midrule

\multirow{1}{*}{ResNet-18~\cite{he2016deep}}
&Supervised Learning & \textbf{0.0818}$\pm$0.0000 & 0.0811$\pm$0.0005 & 0.1743$\pm$0.0030 & 0.0928$\pm$0.0309 & 0.0851$\pm$0.0308 & 0.0782$\pm$0.0176 & 0.0534$\pm$0.0070 \\
\midrule

\multirow{1}{*}{SSL~\cite{ciga2022self}}
&Supervised Learning & 0.0286$\pm$0.0000 & 0.0275$\pm$0.0002 & 0.0669$\pm$0.0189 & 0.0261$\pm$0.0058 & 0.0216$\pm$0.0043 & 0.0276$\pm$0.0077 & 0.0254$\pm$0.0070 \\

\bottomrule
\end{tabular}
\end{adjustbox}

\caption{\textbf{ECE and boundary margin on Vanderbilt.} ECE (Top-1) and boundary margin mean scores (Mean$\pm$SD) across shots $\{0,1,2,4,8,16,32\}$. Column-wise best values are bolded (lower is better for ECE; higher is better for boundary margin).}
\label{tab:ece_margin_vandy}
\end{table*}

\begin{table*}[!t]
\centering
\begin{adjustbox}{max width=\textwidth}
\begin{tabular}{llcccccccccccccc}
\toprule
\multirow{2}{0.4in}{Model} & Method & \multicolumn{7}{c}{Intra-Class Distance} & \multicolumn{7}{c}{Silhouette Score} \\[0.3em]
\cmidrule{3-9}
\cmidrule{10-16}
& Shots & 0 & 1 & 2 & 4 & 8 & 16 & 32 & 0 & 1 & 2 & 4 & 8 & 16 & 32 \\
\midrule

\multirow{4}{*}{CLIP~\cite{radford2021learning}}
& Adapter~\cite{gao2024clip} & 2.4820 & 2.4937 & 2.5021 & 2.3861 & 2.5315 & 2.7667 & 2.7161 & -0.0291 & -0.0342 & -0.0334 & -0.0379 & -0.0365 & -0.0326 & -0.0364 \\
& Classifier~\cite{ioffe2015batch} & 2.4820 & 3.1019 & 3.0256 & 2.7763 & 2.8256 & 2.8579 & 2.7962 & -0.0291 & 0.0044 & 0.0468 & 0.0548 & 0.0087 & 0.0488 & 0.0484 \\
& LoRA~\cite{hu2022lora} & 2.4820 & 2.5154 & 2.5903 & 2.5607 & 2.5249 & 2.3678 & 2.3518 & -0.0291 & -0.0329 & -0.0370 & -0.0402 & -0.0329 & -0.0051 & -0.0155 \\
& Vanilla~\cite{srinivasan2024comparative} & 2.4820 & 2.6004 & 2.6157 & 2.5036 & 2.5971 & 2.5205 & 2.3991 & -0.0291 & -0.0336 & -0.0312 & -0.0332 & -0.0176 & 0.0039 & 0.0300 \\
\midrule

\multirow{4}{*}{CONCH~\cite{lu2024visual}}
& Adapter~\cite{gao2024clip} & \textbf{2.3550} & 2.3776 & \textbf{2.3699} & 2.3708 & 2.3757 & 2.3952 & 2.3008 & \textbf{0.1032} & 0.1045 & 0.1047 & 0.0801 & 0.0998 & 0.0403 & 0.1108 \\
& Classifier~\cite{ioffe2015batch} & \textbf{2.3550} & 2.8274 & 2.8359 & 2.7550 & 2.7543 & 2.7021 & 2.7898 & \textbf{0.1032} & \textbf{0.1917} & \textbf{0.1596} & \textbf{0.1622} & \textbf{0.1811} & \textbf{0.1547} & \textbf{0.1842} \\
& LoRA~\cite{hu2022lora} & \textbf{2.3550} & \textbf{2.3608} & 2.3951 & 2.3772 & 2.3838 & 2.3512 & 2.3034 & \textbf{0.1032} & 0.1026 & 0.1016 & 0.1007 & 0.0959 & 0.0933 & 0.0790 \\
& Vanilla~\cite{srinivasan2024comparative} & \textbf{2.3550} & 2.3704 & 2.3716 & \textbf{2.3556} & \textbf{2.3633} & \textbf{2.3199} & \textbf{2.2636} & \textbf{0.1032} & 0.0982 & 0.0994 & 0.0972 & 0.0981 & 0.0978 & 0.1125 \\
\midrule

\multirow{4}{*}{PLIP~\cite{huang2023visual}}
& Adapter~\cite{gao2024clip} & 3.2031 & 3.1782 & 3.1748 & 3.2420 & 3.2093 & 3.2110 & 3.1746 & 0.0316 & 0.0315 & 0.0288 & 0.0314 & 0.0315 & 0.0314 & 0.0301 \\
& Classifier~\cite{ioffe2015batch} & 3.2031 & 3.1665 & 3.2837 & 3.3406 & 3.2480 & 3.6128 & 3.2242 & 0.0316 & 0.0373 & 0.0566 & 0.0700 & 0.0724 & 0.0546 & 0.0740 \\
& LoRA~\cite{hu2022lora} & 3.2031 & 3.2644 & 3.1885 & 3.2145 & 3.1509 & 3.1498 & 2.9115 & 0.0316 & 0.0316 & 0.0322 & 0.0315 & 0.0321 & 0.0316 & 0.0625 \\
& Vanilla~\cite{srinivasan2024comparative} & 3.2031 & 3.2055 & 3.2171 & 3.1875 & 3.2082 & 3.2044 & 3.0516 & 0.0316 & 0.0312 & 0.0302 & 0.0294 & 0.0439 & 0.0651 & 0.0837 \\
\midrule

\bottomrule
\end{tabular}
\end{adjustbox}
\caption{\textbf{Intra-Class Distance and Silhouette Score on Vanderbilt.}
Few-shot geometric metrics under three backbones: CLIP, CONCH, and PLIP across shot settings $\{0,1,2,4,8,16,32\}$ and four fine-tuning strategies.
Intra-Class Distance measures within-class compactness in the UMAP space (lower is better), and Silhouette Score measures cluster separability in the original embedding space using cosine distance (higher is better). For Classifier tuning, both metrics are computed from the penultimate representation of the tuned MLP head.}
\label{tab:distance_vandy_run9}
\end{table*}

\begin{table*}[ht]
\centering

\begin{adjustbox}{max width=\textwidth}
\begin{tabular}{llc}
\toprule
Model & Method & Samples/s ($\mu \pm \sigma$) \\
\midrule

\multirow{5}{*}{CLIP~\cite{radford2021learning}}
&0-shot & 38.8668 $\pm$ 0.0000 \\
&Adapter~\cite{gao2024clip} & 39.1536 $\pm$ 0.0325 \\
&Classifier~\cite{ioffe2015batch} & 41.5737 $\pm$ 0.4636 \\
&LoRA~\cite{hu2022lora} & 41.2589 $\pm$ 0.4928 \\
&Vanilla~\cite{srinivasan2024comparative} & 40.7104 $\pm$ 0.6712 \\
\midrule

\multirow{5}{*}{CONCH~\cite{lu2024visual}}
&0-shot & 29.3611 $\pm$ 0.0000 \\
&Adapter~\cite{gao2024clip} & 30.6423 $\pm$ 0.2150 \\
&Classifier~\cite{ioffe2015batch} & 31.4967 $\pm$ 0.0799 \\
&LoRA~\cite{hu2022lora} & 29.5964 $\pm$ 0.2099 \\
&Vanilla~\cite{srinivasan2024comparative} & 30.6389 $\pm$ 0.4046 \\
\midrule

\multirow{5}{*}{PLIP~\cite{huang2023visual}}
&0-shot & 42.3505 $\pm$ 0.0000 \\
&Adapter~\cite{gao2024clip} & 43.6870 $\pm$ 0.4841 \\
&Classifier~\cite{ioffe2015batch} & 42.4355 $\pm$ 0.5230 \\
&LoRA~\cite{hu2022lora} & 43.4285 $\pm$ 0.7742 \\
&Vanilla~\cite{srinivasan2024comparative} & 42.6651 $\pm$ 0.6393 \\
\midrule

ResNet-18~\cite{he2016deep} & Supervised Learning & 47.0729 $\pm$ 1.2160 \\
\midrule

SSL~\cite{ciga2022self} & Supervised Learning & \textbf{52.5185 $\pm$ 3.5419} \\

\bottomrule
\end{tabular}
\end{adjustbox}
\caption{\textbf{Inference throughput on Cornell.} Samples processed per second during inference on a 1,000-sample timing subset, summarized as mean$\pm$SD. For VLMs, the 0-shot row reports base model inference timing, while Adapter, Classifier, LoRA, and Vanilla rows summarize shots $\{1,2,4,8,16,32\}$. ResNet-18 and SSL rows summarize shots $\{0,1,2,4,8,16,32\}$. Higher values indicate faster inference; bold indicates the highest mean throughput.}
\label{tab:timing_cornell}
\end{table*}

\begin{table*}[ht]
\centering

\begin{adjustbox}{max width=\textwidth}
\begin{tabular}{llc}
\toprule
Model & Method & Samples/s ($\mu \pm \sigma$) \\
\midrule

\multirow{5}{*}{CLIP~\cite{radford2021learning}}
&0-shot & 32.6514 $\pm$ 0.0000 \\
&Adapter~\cite{gao2024clip} & 35.9904 $\pm$ 5.9719 \\
&Classifier~\cite{ioffe2015batch} & 39.5126 $\pm$ 0.4868 \\
&LoRA~\cite{hu2022lora} & 38.7514 $\pm$ 0.5405 \\
&Vanilla~\cite{srinivasan2024comparative} & 36.5264 $\pm$ 6.3853 \\
\midrule

\multirow{5}{*}{CONCH~\cite{lu2024visual}}
&0-shot & 27.6750 $\pm$ 0.0000 \\
&Adapter~\cite{gao2024clip} & 28.3857 $\pm$ 0.6497 \\
&Classifier~\cite{ioffe2015batch} & 31.2421 $\pm$ 0.3824 \\
&LoRA~\cite{hu2022lora} & 29.0971 $\pm$ 0.6714 \\
&Vanilla~\cite{srinivasan2024comparative} & 27.4787 $\pm$ 4.3375 \\
\midrule

\multirow{5}{*}{PLIP~\cite{huang2023visual}}
&0-shot & 39.7531 $\pm$ 0.0000 \\
&Adapter~\cite{gao2024clip} & 41.6119 $\pm$ 0.7270 \\
&Classifier~\cite{ioffe2015batch} & 40.0198 $\pm$ 0.9968 \\
&LoRA~\cite{hu2022lora} & 40.1168 $\pm$ 1.6200 \\
&Vanilla~\cite{srinivasan2024comparative} & 39.8413 $\pm$ 5.0300 \\
\midrule

ResNet-18~\cite{he2016deep} & Supervised Learning & 43.1835 $\pm$ 0.9697 \\
\midrule

SSL~\cite{ciga2022self} & Supervised Learning & \textbf{45.8984 $\pm$ 1.8326} \\

\bottomrule
\end{tabular}
\end{adjustbox}
\caption{\textbf{Inference throughput on Vanderbilt.} Samples processed per second during inference on a 1,000-sample timing subset, summarized as mean$\pm$SD. For VLMs, the 0-shot row reports base model inference timing, while Adapter, Classifier, LoRA, and Vanilla rows summarize shots $\{1,2,4,8,16,32\}$. ResNet-18 and SSL rows summarize shots $\{0,1,2,4,8,16,32\}$. Higher values indicate faster inference; bold indicates the highest mean throughput.}
\label{tab:timing_vandy}
\end{table*}

\subsection{Result Analysis on Vanderbilt}
\paragraph{Fine-tuning strategy.}
Across Vanderbilt shot settings, adaptation strategy is the dominant source of performance variation (Table~\ref{tab:table_2}; Fig.~\ref{fig:vandy_roc_auc}). Vanilla fine-tuning provides the most consistent gains from low to high supervision, with CONCH with Vanilla reaching the strongest overall high-shot endpoint. Adapter and LoRA are generally intermediate and remain competitive in specific regimes, but their trajectories are less uniformly strong across all metrics. By contrast, Classifier tuning is the least stable strategy on Vanderbilt and frequently degrades as shot increases, including severe collapse for CONCH and CLIP in both Accuracy and F1.

\paragraph{Backbone domain knowledge.}
Backbone pretraining domain relevance is clearly reflected in both low-shot and high-shot behavior. In 0-shot AUC, pathology-specialized backbones substantially exceed general CLIP, indicating a stronger pathology prior before adaptation. With additional supervision, CONCH preserves the highest ceiling across metrics, while PLIP remains competitive but typically below CONCH at medium-to-high shots. CLIP improves with supervision, especially under Vanilla/LoRA, but remains behind pathology-native backbones in the strongest Vanderbilt settings. This pattern is consistent with a domain-knowledge effect that improves both initialization quality and attainable high-shot performance.

\paragraph{Cross-modal indicators.}
Cross-modal trajectories follow the expected direction with increasing supervision (Fig.~\ref{fig:vandy_alignment_sim_gap}): Alignment tends to rise, and Similarity Gap moves from near-zero toward positive values. UMAP/KDE visualizations (Fig.~\ref{fig:vandy_kde}) and geometric summaries (Table~\ref{tab:distance_vandy_run9}) show progressively clearer organization at higher shots, especially for CONCH with Vanilla and CONCH with Adapter. However, compared with the Cornell trend strength, Vanderbilt shows flatter curves and more method overlap in several settings, indicating weaker separation in representation quality despite directional improvement.

\paragraph{Calibration and boundary margin.}
Confidence-related behavior is more heterogeneous than discrimination metrics on Vanderbilt (Table~\ref{tab:ece_margin_vandy}). Even when AUC and Accuracy improve, ECE often remains high at medium and high shots, and Classifier tuning shows pronounced calibration failure in high-shot settings. Boundary margin generally increases with supervision for stronger methods, but high margin does not always correspond to high correctness. This mismatch indicates persistent confident-error risk under Vanderbilt domain conditions.

\subsection{Cross-Center Comparison}
Comparing Cornell and Vanderbilt reveals shared directional trends but different effect magnitudes. In both centers, performance improves with shot count, pathology-specialized backbones outperform general CLIP, and Vanilla is usually the most reliable adaptation strategy at medium-to-high shots. However, Vanderbilt exhibits stronger method sensitivity and weaker absolute discrimination endpoints, especially in Accuracy and F1, while Cornell shows steeper and more stable gains across multiple backbone-strategy pairs.

The largest center-level divergence is Classifier tuning. On Cornell, Classifier tuning can remain competitive for CONCH (including strong AUC at higher shots), whereas on Vanderbilt it degrades substantially across backbones and metrics, with poor robustness as supervision grows. A second divergence appears in confidence quality: Cornell high-shot settings more often achieve jointly favorable discrimination and calibration, while Vanderbilt retains higher ECE in many high-performing configurations. Taken together, these results support a consistent cross-center conclusion: trend direction is reproducible, but practical reliability remains center-dependent and requires strategy-specific validation under external distribution shift.


\clearpage
\bibliography{sample}





\end{document}